\documentclass[10pt,journal,compsoc]{IEEEtran}

\usepackage{amsmath,setspace}
\usepackage{graphicx}
\usepackage{subcaption}
\usepackage{amssymb}
\usepackage{color}
\usepackage{bm}
\usepackage{bbm}
\usepackage[table]{xcolor}
\usepackage{url}
\usepackage{varwidth}
\usepackage{gensymb}

\ifCLASSOPTIONcompsoc
  \usepackage[nocompress]{cite}
\else
  \usepackage{cite}
\fi

\DeclareMathOperator*{\argmin}{\arg\!\min}
\DeclareMathOperator*{\argmax}{\arg\!\max}

\DeclareMathOperator{\E}{\mathbb{E}}

\title{Joint Multi-Dimensional Model for Global and Time-Series Annotations}

\author{Anil~Ramakrishna,~\IEEEmembership{Student Member,~IEEE,}
        Rahul~Gupta,~\IEEEmembership{Member,~IEEE,}
        and~Shrikanth~Narayanan,~\IEEEmembership{Fellow,~IEEE}%
}

\makeatletter
\def\name#1{\gdef\@name{#1\\}}

\begin{document}

\IEEEtitleabstractindextext{%
\begin{abstract}
Crowdsourcing is a popular approach to collect annotations for unlabeled data instances. It involves collecting a large number of annotations from several, often naive untrained annotators for each data instance which are then combined to estimate the ground truth. Further, annotations for constructs such as affect are often multi-dimensional with annotators rating multiple dimensions, such as valence and arousal, for each instance. Most annotation fusion schemes however ignore this aspect and model each dimension separately. In this work we address this by proposing a generative model for multi-dimensional annotation fusion, which models the dimensions jointly leading to more accurate ground truth estimates. The model we propose is applicable to both global and time series annotation fusion problems and treats the ground truth as a latent variable distorted by the annotators. The model parameters are estimated using the Expectation-Maximization algorithm and we evaluate its performance using synthetic data and real emotion corpora as well as on an artificial task with human annotations. 
\end{abstract}

}

\maketitle

\section{Introduction}
\label{sec:intro}
Crowdsourcing is a popular tool used in collecting human judgments on subjective constructs such as emotion. Typical examples include annotations of images and video clips with categorical emotions or with continuous affective dimensions such as \textit{valence} or \textit{arousal}. Online platforms such as Amazon Mechanical Turk\footnote{www.mturk.com} (MTurk) and Crowdflower\footnote{www.crowdflower.com} have risen in popularity owing to their inexpensive annotation costs and their ability to scale efficiently.

Crowdsourcing is also a popular approach in collecting labels for training supervised machine learning algorithms. Such labels are typically obtained from domain experts, which can be slow and expensive. For example, in the medical domain, it is often expensive to collect diagnosis information given laboratory tests since this requires judgments from trained professionals. On the other hand, unlabeled patient data may be easily available. Crowdsourcing has been particularly successful in such settings with easy availability of unlabeled data instances since we can collect a large number of annotations from untrained and inexpensive workers over the Internet, which when combined together may be comparable or even better than expert annotations \cite{surowiecki2005wisdom}.

A typical crowdsourcing setting involves collecting annotations from a large number of workers; hence there is a need to robustly combine them to estimate the ground truth. The most common approach for this is to take simple averages for continuous annotations or perform majority voting for categorical annotations. However, this assumes uniform competency across all the workers which is not always guaranteed or justified. Several alternative approaches have been proposed to address this challenge, each assuming a specific function modeling the annotators' behavior. In practice, it is common to collect annotations on multiple questions for each data instance in order to reduce costs, the annotators' mental load or even to improve annotation accuracy. For example, if we're annotating valence and arousal for a given data instance (such as a single image or video segment), collecting annotations on both these dimensions in one session per instance may be preferred over collecting valence annotations for all instances followed by arousal. 

Such a joint annotation task may entail \textit{task specific} or \textit{annotator specific} dependencies between the annotated dimensions. In the aforementioned example, task specific dependencies may occur due to inherent correlations between the valence and arousal dimensions depending on the experimental setup. Annotator specific dependencies may occur due to a given annotator's (possibly incorrect or incomplete) understanding of the annotation dimensions. Hence it is of relevance to model the dimensions jointly. However, most state of the art models in annotation fusion combine the annotations by treating the different dimensions independently. 

Joint modeling of the annotation dimensions may result in more accurate estimates of the ground truth as well as in giving a better picture of the annotators' behavior. In this work, we address this goal by proposing a multi-dimensional model which makes use of any potential relationships between the annotation dimensions while combining them. The model we propose is applicable to both the global annotation setting (such as while collecting emotion annotations on a picture, judgment about the overall tone of a conversation, etc.) as well as time series annotations (for example, time continuous annotations of audio/video clips on dimensions such as engagement or affect). Our model treats the hidden ground truth as latent variables and estimates them jointly with the annotator parameters using the Expectation Maximization (EM) algorithm \cite{dempster1977maximum}. We evaluate the model in both settings with both synthetic and real emotion corpora. We also create an artificial annotation task with controlled ground truth which is used in the model evaluation for both settings.

The main contributions of this work are as follows:

\begin{enumerate}
\item We propose a unified model to capture relationships between annotation dimensions. For ease of exposition we focus on the linear case in this paper.
\item The linear model we propose results in an annotator specific matrix which captures this annotator level relationship between the annotation dimensions. 
\item We create a novel multi-dimensional annotation task with controlled ground truth and use it to evaluate both the global and time series annotation settings of the model.
\end{enumerate}

The rest of the paper is organized as follows. In Section \ref{sec:related}, we review related work and motivate the problem in Section \ref{sec:motivation}. In Section \ref{sec:model}, we describe the proposed model and provide equations for parameter estimation using EM algorithm (derivations are deferred to the appendix). We evaluate the model in Section \ref{sec:exp} and provide conclusions in Section \ref{sec:conc}.

\section{Related work}
\label{sec:related}
Several authors, most notably \cite{surowiecki2005wisdom}, assert the benefits of aggregating opinions from many people which is often believed to be better than those from a small number of experts, under certain conditions. Often referred to as the \textit{wisdom of crowds}, this approach has been remarkably popular in recent times, specially in fields such as psychology and behavioral sciences where a ground truth may not be easily accessible or may not exist. This popularity can be largely attributed to online crowdsourcing platforms such as Mturk that connect researchers with low cost workers from around the globe. Along with cost, scalability is another major appeal with such tools leading to their frequent use in machine learning, leveraging large scale annotation of data instances such as images \cite{deng2009imagenet}, audio/video clips \cite{vondrick2013efficiently} and text snippets \cite{snow2008cheap}. 

\begin{figure}
\centering
\includegraphics[trim={0.75cm 0cm 0cm 0cm},scale=0.4]{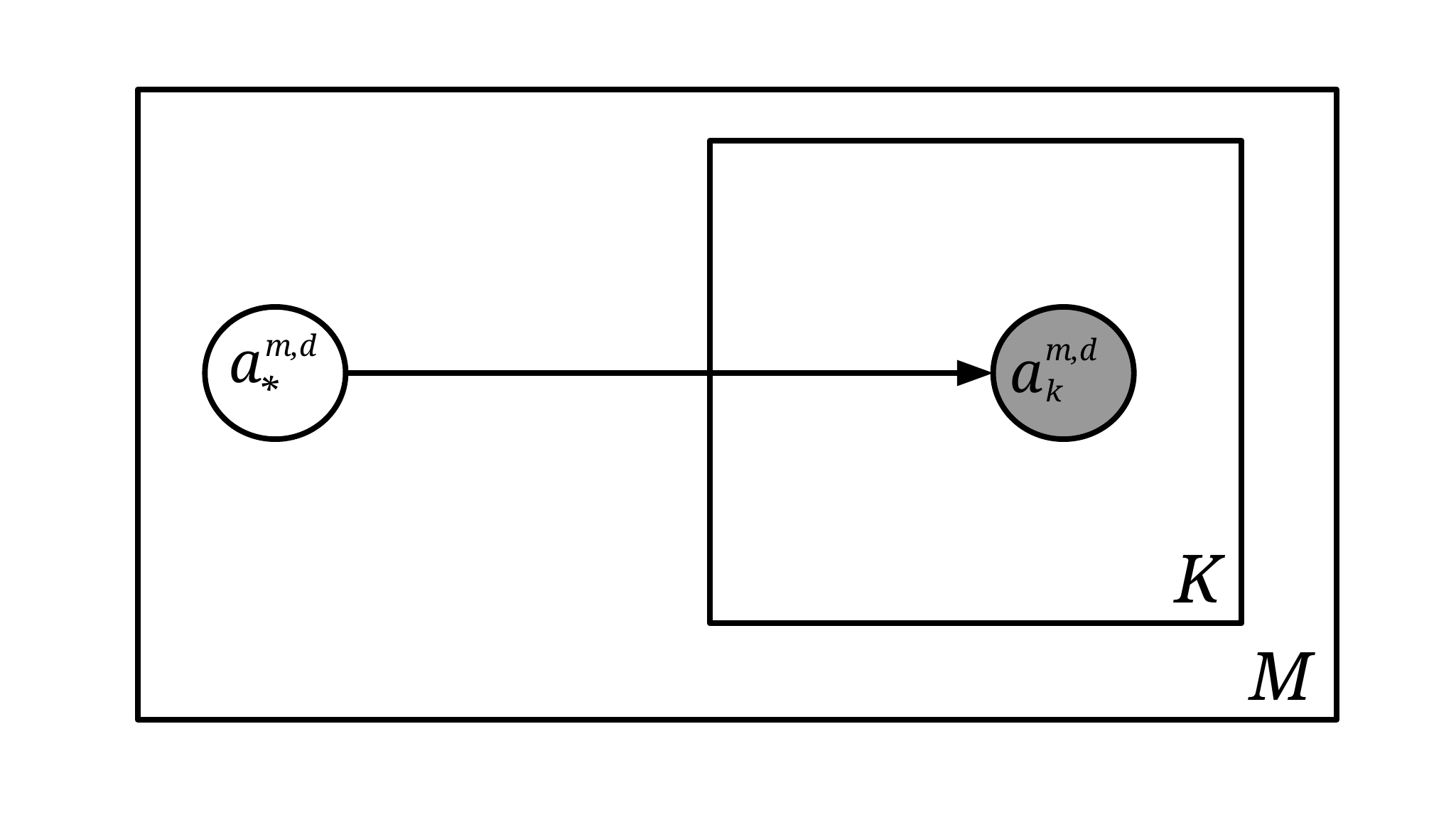}
\vspace{-20pt}
\caption{Plate notation for a \underline{basic annotation model}. $a^{m,d}_*$ is the latent ground truth for the given data instance (for the $d^\text{th}$ question) and $a^{m,d}_k$ is the rating provided by the $k^\text{th}$ annotator.} 
\label{fig:modela}
\vspace{-5pt}
\end{figure}

Figure \ref{fig:modela} shows a common setting in the crowdsourcing paradigm. For each data instance $m$, annotator $k$ provides a noisy annotation $a^{m,d}_k$ which depends on the ground truth $a^{m,d}_*$ where $d$ is the dimension being annotated. Since we collect several annotations for each $m$, we need to aggregate them to estimate the unknown ground truth. The most common technique used in this aggregation is to take the average value in case of numeric annotations or perform majority voting in the case of categorical annotations as shown in Equation \ref{eqn:model0eqn}. 

\begin{equation}
a^{m,d}_* = \argmax_j \sum_k \mathbbm{1}\{ a^{m,d}_k == j \}
\label{eqn:model0eqn}
\end{equation}

\noindent where, $\mathbbm{1}$\{\} is the indicator function.

While simple and easy to implement, this approach assumes consistent reliability among the different annotators which seems unreasonable, especially in online platforms such as Mturk. To address this, several approaches have been suggested that account for annotator reliability in estimating the ground truth.

Early efforts to capture reliability in annotation modeling \cite{dawid1979maximum}, \cite{smyth1995inferring} assumed specific structure to the functions modeled by each annotator. Given a set of annotations $a^{m,d}_k$ along with the corresponding function parameters, the ground truth is estimated using the Maximum A Posteriori (MAP) estimator.

\begin{equation}
a^{m,d}_* = \argmax_j \sum_k \log p(a^{m,d}_k|a^{m,d}_* = j) + \log p(a^{m,d}_* = j)
\end{equation}

\noindent where $p(a^{m,d}_*)$ is the prior probability of ground truth.

In \cite{dawid1979maximum}, the categorical ground truth label $a^{m,d}_* = i$ is modified probabilistically by annotator $k$ using a stochastic matrix $\Pi_k$ as shown in Equation \ref{eqn:dawid} in which each row is a multinomial conditional distribution given the ground truth.

\begin{equation}
P(a^{m,d}_k = j | a^{m,d}_* = i) = \pi^k_{ij}
\label{eqn:dawid}
\end{equation}

Given annotations from $K$ different annotators, their parameters $\Pi_k$ and prior distribution of labels $p_j = P(a^{m,d}_* = j)$, the ground truth is estimated using MAP estimation as before.

\begin{equation}
a^{m,d}_* = \argmax_{j} \sum_{k} \log \pi_{j({a^{m,d}_k})} + \log p_j
\end{equation}

The above expression makes a conditional independence assumption for annotations given the ground truth label. 
Since we do not typically have the annotator parameters $\Pi^k$, these are estimated using the EM algorithm. 

\begin{figure}
\centering
\includegraphics[trim={0.7cm 0.75cm 0cm 0cm},scale=0.45]{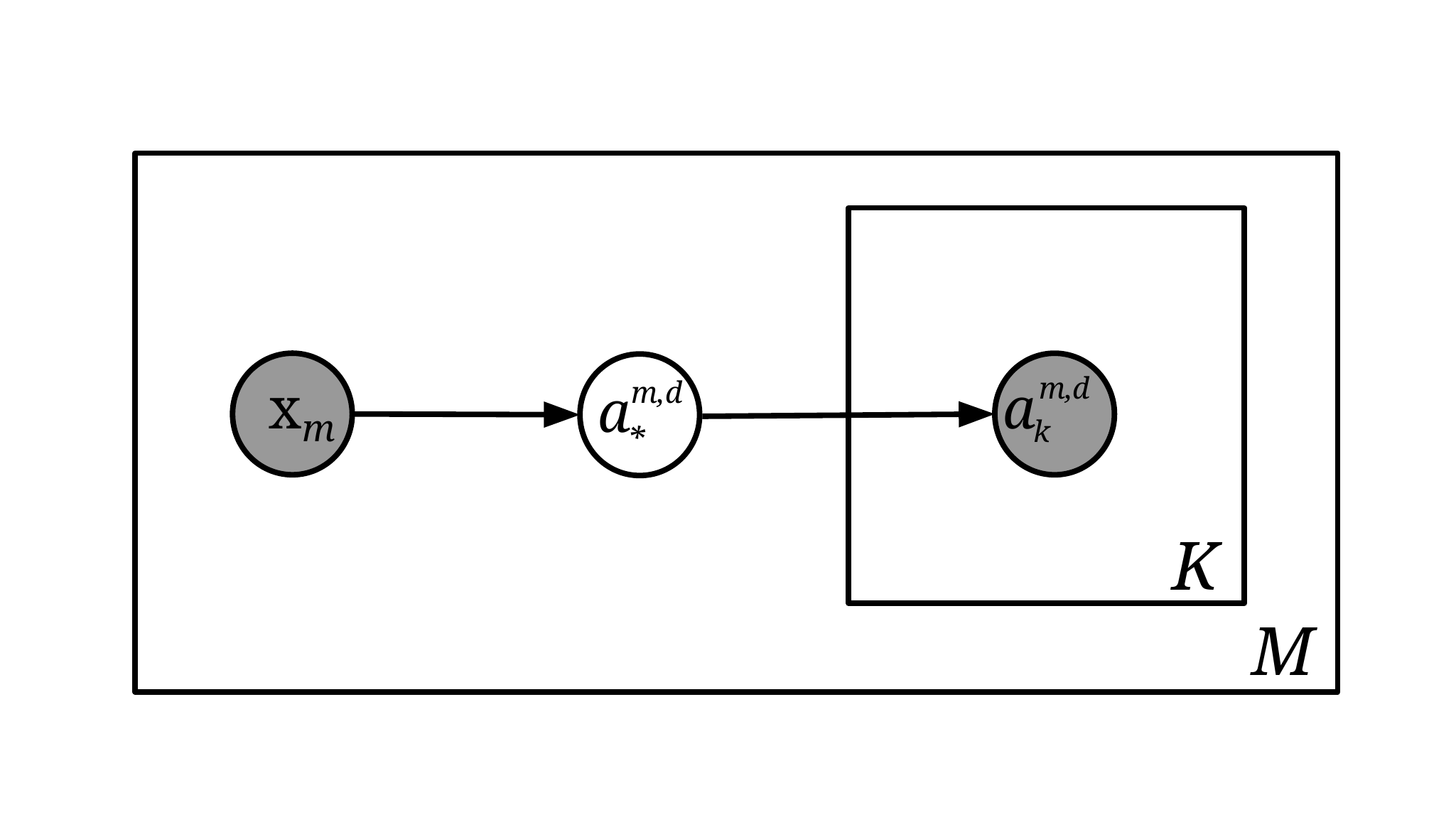}
\vspace{-20pt}
\caption{Annotation model proposed by \cite{raykar2010learning} with a jointly learned predictor. $\textbf{x}_m$ is the set of features for the $m^\text{th}$ data instance; $a^{m,d}_*$ is the $d^{th}$ dimension of the latent ground truth which is modeled as a function of $\textbf{x}_m$; $a^{m,d}_k$ is the rating provided by the $k^\text{th}$ annotator. 
}
\vspace{-5pt}
\label{fig:modelb}
\end{figure}

Figure \ref{fig:modelb} shows an extension of the model in Figure \ref{fig:modela} in which we learn a predictor (classifier/regression model) for the ground truth jointly with annotator parameters. Such a predictor may be used to estimate the ground truth for new unlabeled data instances. This strategy of jointly modeling the annotator functions as well as the ground truth predictor has been shown to have better performance when compared to predictors trained independently using the estimated ground truth \cite{raykar2010learning}. The ground truth estimate in this model is given by

\begin{equation}
a^{m,d}_* = \argmax_{a^{m,d}_*} \sum_k \log p(a^{m,d}_k|a^{m,d}_*) + \log p(a^{m,d}_*|\textbf{x}_m)
\label{eqn:modelbeqn}
\end{equation}

Recently, several additional extensions have been proposed to the model in Figure \ref{fig:modelb}; For example, in \cite{audhkhasi2013globally}, the authors assume varying regions of annotator expertise in the data feature space and account for this using different probabilities for label confusion for each region. The authors show that this leads to a better estimation of annotator reliability and ground truth. 

The models described so far have been designed for annotation tasks in which the task is to rate some global property of the data. For example, in image based emotion annotation, the task may be to provide annotations on affective dimensions such as valence and arousal conveyed by each image. However, human interactions often involve variations of these dimensions over time \cite{Metallinou2013AnnotationandProcessingof} which are captured using time series annotations from audio/video clips. Various tools have been developed to collect such annotations, including Anvil \cite{kipp2001anvil}, Feeltrace \cite{cowie2000feeltrace}, EMuJoy \cite{nagel2007emujoy}, Gtrace \cite{cowie2011gtrace} and DARMA \cite{girard2018darma} (for a review of available tools and their properties, see \cite{dupre2015oudjat} and \cite{girard2018darma}). In fusing such time series annotations, the previously mentioned models are applicable only if annotations from each frame are treated independently. However, this entails several unrealistic assumptions such as independence between frames, zero lag in the annotators and synchronized response in the annotators to the underlying stimulus. 

Several works have been proposed to capture the underlying reaction lag in the annotators. \cite{nicolaou2014dynamic} proposed a generalization of Probabilistic Canonical Correlation Analysis (PCCA) \cite{bach2005probabilistic} named Dynamic PCCA which captures temporal dependencies of the shared ground truth space in a generative setting, and incorporated a latent time warping process to implicitly handle the reaction lags in annotators. They have further proposed a supervised extension of their model which jointly learns a predictor function for the latent ground truth signal similar to \cite{raykar2010learning}. \cite{mariooryad2015correcting} address the reaction lag by explicitly finding the time shift that maximizes the mutual information between expressive behaviors and their annotations. \cite{Gupta2015} generalize the work of \cite{mariooryad2015correcting} by  using a linear time invariant (LTI) filter which can also handle any bias or scaling the annotators may introduce. 

More recent works in annotation fusion include \cite{kara2015modeling} in which the authors propose a variant of the model in Figure \ref{fig:modela} with various annotator functions to capture four specific types of annotator behavior. \cite{shah2015approval} describes a mechanism named approval voting that allows annotators to provide multiple answers instead of one for instances where they are not confident.
\cite{sheng2008get} uses repeated sampling for opinions from annotators over the same data instances to increase reliability in annotations.

Most of the models described above focus on combining annotations on each dimension separately. However, the annotation dimensions are often related. For example, many studies in emotion literature have reported interrelationships between discrete emotion categories \cite{watson1992traits, russell1999bipolarity}. The circumplex model \cite{russell1980circumplex}, which attempts to capture these relationships by modeling the emotions as points on a two dimensional space, has also been noted to exhibit \textit{v-shaped} patterns in the joint distribution of valence and arousal \cite{malandrakis2011supervised}.
In addition, in most practical applications, the annotation tasks themselves are multi-dimensional. For example, while collecting ratings on affective dimensions it is routine to collect annotations on valence, arousal and dominance together. Further, there may be dependencies between the internal definitions the annotators hold for the annotation dimensions; for example, while annotating emotional dimensions, a particular annotator may associate certain valence values with only a certain range of arousal. Hence it may be beneficial to model the different dimensions jointly while performing annotation fusion. However, research in this direction has been limited. \cite{Ramakrishna2016} proposed a model which assumes joint Gaussian noise between the annotation dimensions, but their model fails to capture structural dependencies described above between the annotation and ground truth dimensions. The model proposed in \cite{nicolaou2014dynamic} can indeed be generalized to combine the different annotation dimensions together but they do not evaluate with joint annotated dimensions from a real dataset as that is not the focus of their work. \cite{sharma2019functional} jointly model continuous annotations on valence and arousal using \textit{personalized basis spline} functions, on which functional PCA is applied to identify the dominant spline functions. Using this model, they estimate the ground truth for each data instance using a heuristic algorithm, but their model does not include a jointly trained ground truth predictor. 
It is therefore of relevance to model multi-dimensional annotation fusion as part of the unified annotator function and predictor modeling paradigm.

In this work, we propose a joint multi-dimensional model to address many of the gaps mentioned above. Our model captures annotator specific linear relationships between different annotation dimensions, and is an extension of the Factor Analysis model \cite{harman1976modern}. It incorporates an annotator specific transformation matrix parameter $F_k$, which explicitly captures the relationship between the annotation dimensions and enables clear interpretations of the estimated relationships; the matrix $F_k$ is jointly estimated with a predictor for the ground truth signal. We further provide generalizations of our model to both global and time series annotation settings. We begin with a motivation followed by a detailed description of the model and its parameter estimation in the next sections. 

\begin{figure*}%
	\centering
    \begin{subfigure}[t]{0.25\textwidth}
        \centering
        \includegraphics[height=1.2in]{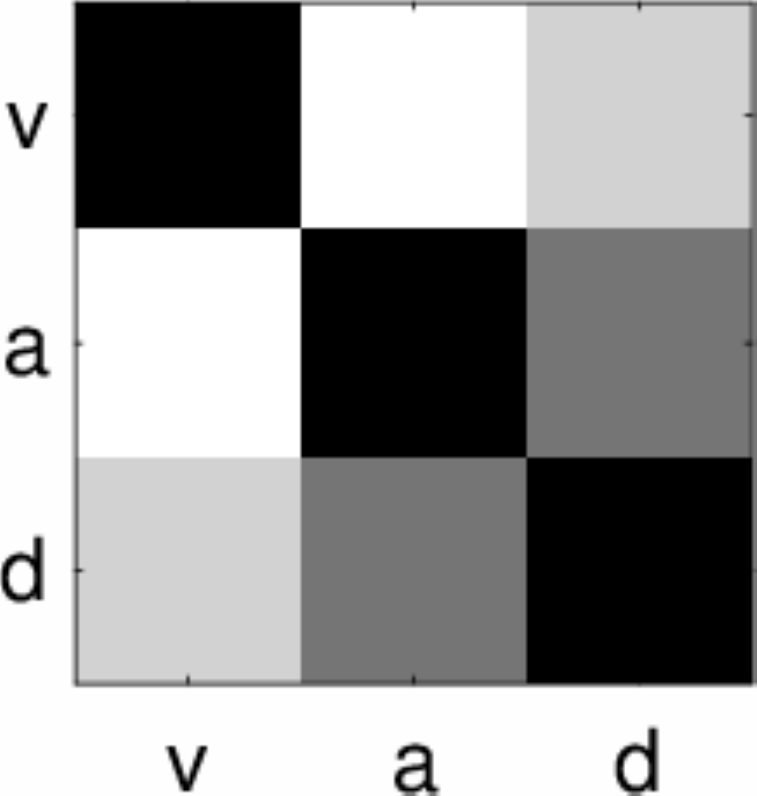}
        \caption{IEMOCAP}
    \end{subfigure}%
    ~ 
    \begin{subfigure}[t]{0.25\textwidth}
        \centering
        \includegraphics[height=1.2in]{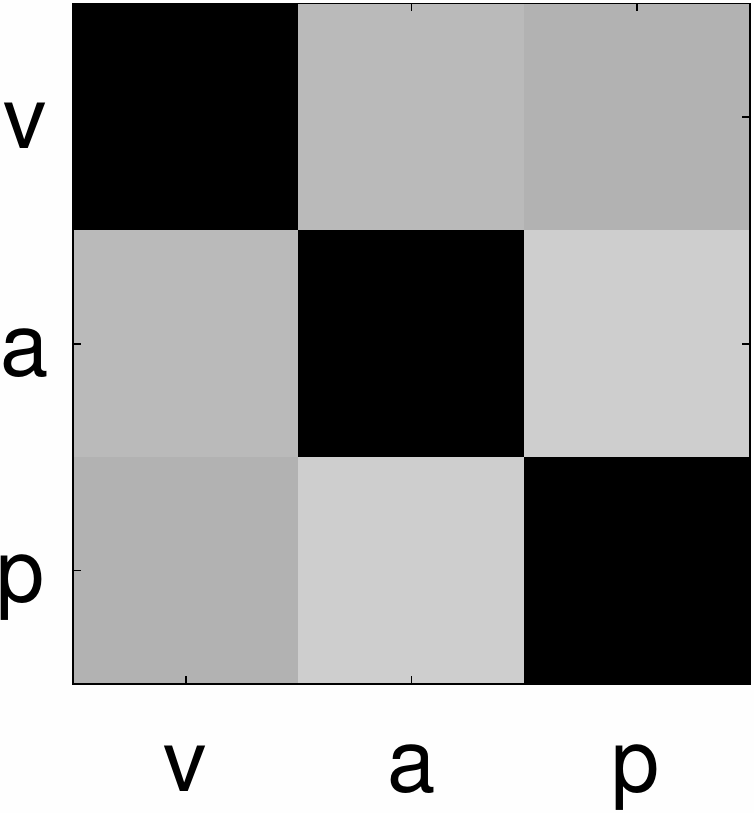}
        \caption{SEMAINE}
    \end{subfigure}%
    ~
    \begin{subfigure}[t]{0.25\textwidth}
        \centering
        \includegraphics[height=1.2in]{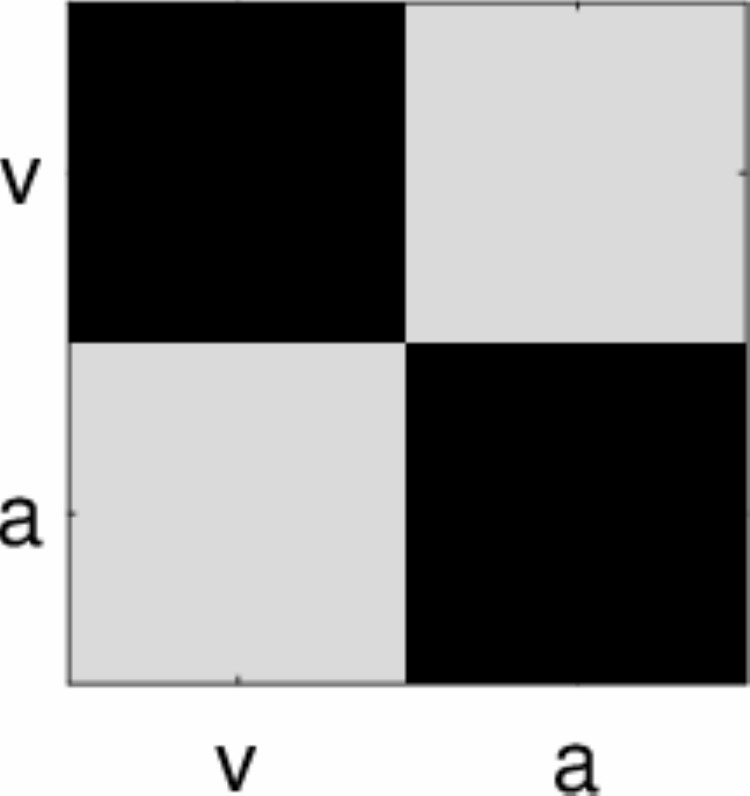}
        \caption{Movie emotions}
    \end{subfigure}%
    ~
    \begin{subfigure}[t]{0.25\textwidth}
        \centering
        \includegraphics[height=1.2in]{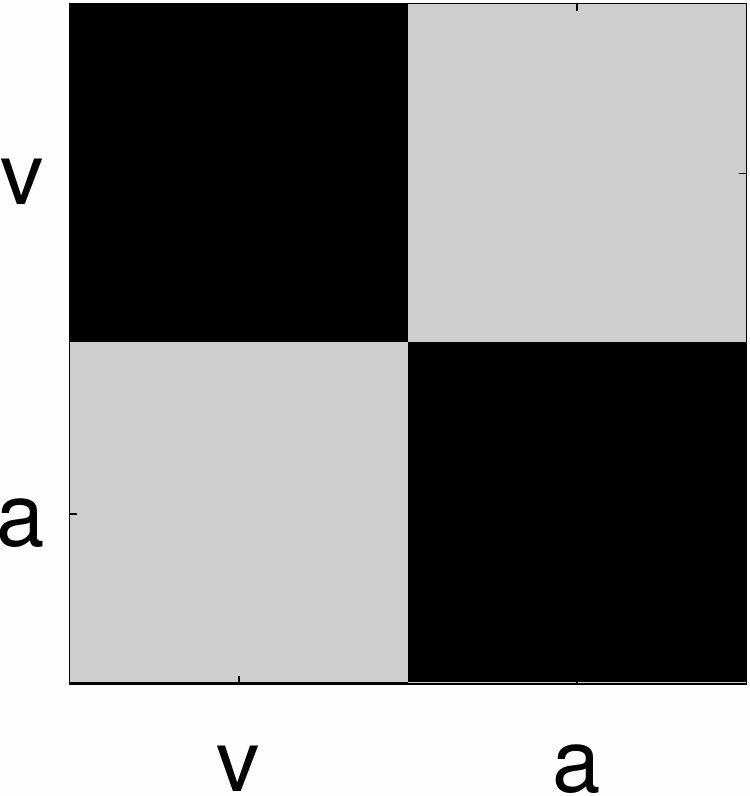}
        \caption{RECOLA}
    \end{subfigure}
    \par\bigskip
    \begin{subfigure}[t]{\textwidth}
    	\centering
    	\includegraphics[]{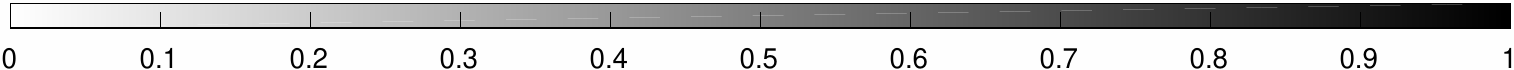}      
    \end{subfigure}
    \caption{Correlation heatmaps for annotations from a representative sample of emotion annotated datasets; v - valence, a - arousal, d - dominance, p - power}
    \label{fig:corrs}
\end{figure*}

\section{Motivation}
\label{sec:motivation}

To examine the relationships between the annotation dimensions, we created a plot of absolute values of Pearson's correlation between annotation dimensions from four commonly studied emotional corpora in Figure \ref{fig:corrs}: IEMOCAP \cite{busso2008iemocap}, SEMAINE \cite{mckeown2012semaine}, RECOLA \cite{ringeval2013introducing} and the movie emotion corpus from \cite{malandrakis2011supervised}. Each of these corpora include annotations over affective dimensions such as valence, arousal, dominance and power. For the IEMOCAP corpus, we used global annotations while the others include time series annotations of the affective dimensions from videos. In each case, the correlations were computed from concatenated annotation values between all the dimensions.

As is evident, in almost all cases, the annotation dimensions exhibit non-zero correlations. 
We attribute the inconsistent correlations between the dimensions across corpora to varying underlying affective narratives as well as differences in perceptions and biases introduced by individual annotators themselves (see Section \ref{appendix_analysis_corrs}). The non-zero correlations highlight the benefit of modeling the annotation dimensions jointly. 
The model we propose is aimed at addressing this. We explain the model in detail in the next section.

\section{Joint Multi-dimensional Annotation Model}
\label{sec:model}

\subsection{Setup}

\begin{figure}
\centering
\includegraphics[trim={0.5cm 0cm 0cm 0cm},scale=0.45]{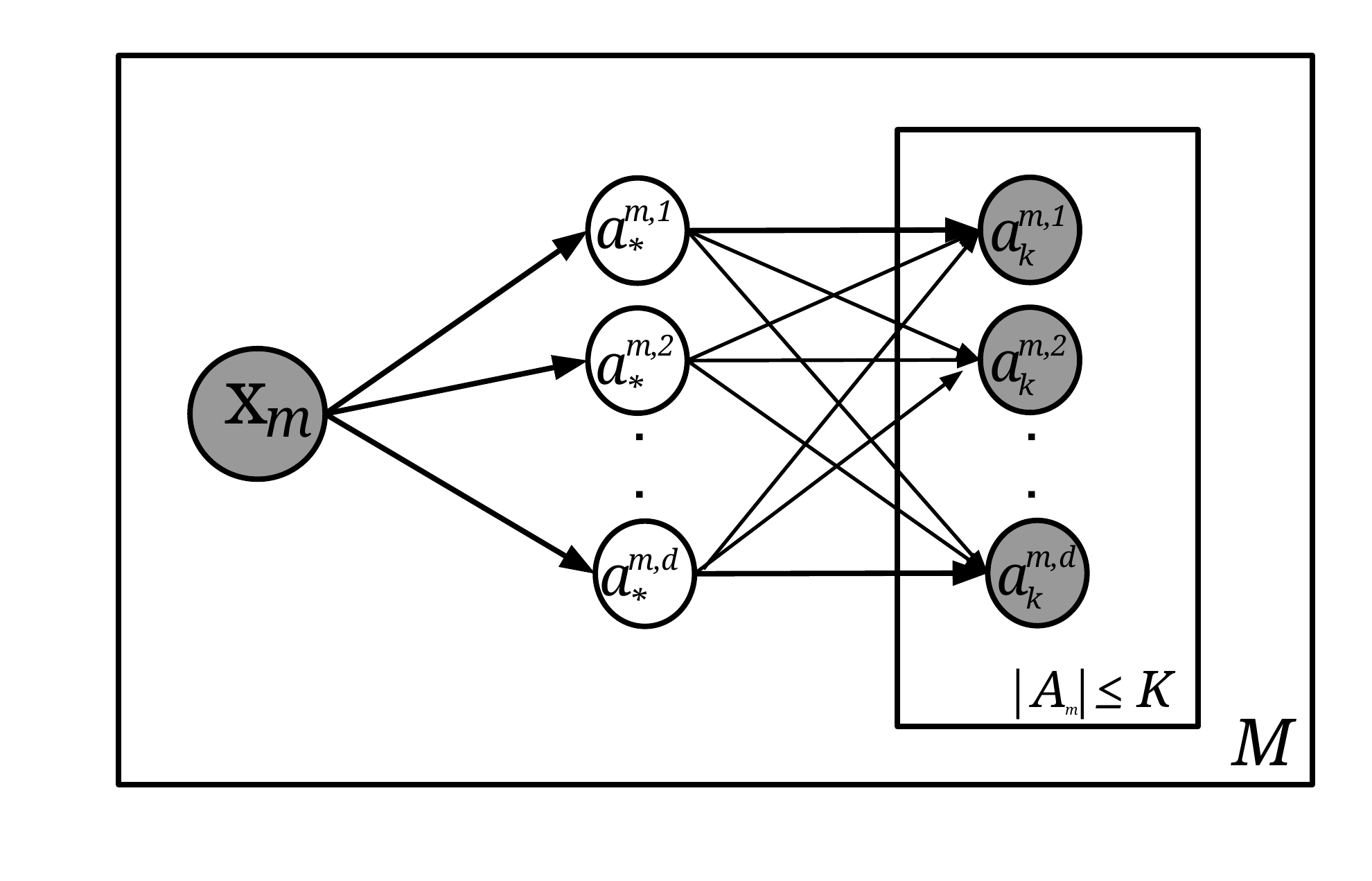}
\vspace{-20pt}
\caption{Proposed model. $\textbf{x}_m$ is the set of features for the $m^\text{th}$ data instance, $a^{m,d}_*$ is the latent ground truth for the $d^{th}$ dimension and $a^{m,d}_k$ is the rating provided by the $k^\text{th}$ annotator. Vectors $\textbf{x}_m$ and $\textbf{a}^m_k$ (shaded) are observed variables, while $\textbf{a}^m_*$ is latent. $\textbf{A}_m$ is the set of annotator ratings for the $m^\text{th}$ instance.}
\vspace{-5pt}
\label{fig:modelc}
\end{figure}

The proposed model is shown in Figure \ref{fig:modelc}. Each data instance $m$ has a feature vector $\textbf{x}_m$ and an associated multidimensional ground truth $\textbf{a}^m_*$, which is defined as follows,
\begin{align}
\textbf{a}^m_* &= f(\textbf{x}_m; \Theta) + \bm{\epsilon}_m 
\end{align}
\label{eqn:gt_eqn}
We assume that from a pool of $K$ annotators, a subset operates on each data instance and provides their annotation $\textbf{a}^m_k$.
\begin{align}
\textbf{a}^m_k &= g(\textbf{a}^m_*; F_k) + \bm{\eta}_k \label{eqn:ann_eqn}
\end{align}

where index $k$ corresponds to the $k^{th}$ annotator; $F_k$ is an annotator specific matrix that defines his/her linear weights for each output dimension; $\bm{\epsilon_m}$ and $\bm{\eta_k}$ are noise terms defined individually in the next sections along with the functions $f$ and $g$. In the global annotation setting, both $\textbf{a}^m_*$ and $\textbf{a}^m_k \in \rm I\!R^D$ where $D$ is the number of items being annotated; for the time series setting $\textbf{a}^m_*$ and $\textbf{a}^m_k$ $ \in \rm I\!R^{T \times D}$, where $T$ is the total duration of the data instance (audio/video signal). 
In all subsequent definitions, we use uppercase letters $M,K,T,D$ to denote various counts and lowercase letters $m,k,t,d$ to denote the corresponding index variables.

We make the following assumptions in our model. 
\begin{enumerate}
\item[A1] Annotations are independent for different data instances. 
\item[A2] The annotations for a given data instance are independent of each other given the ground truth. 
\item[A3] The model ground truths for different annotation dimensions are assumed to be conditionally independent of each other given the features $\bm{x}_m$.
\end{enumerate}

\subsection{Global annotation model}
\label{subsec:disc_model}
In this setting, the ground truth and annotations are $d$ dimensional vectors for each data instance. We define the ground truth $\textbf{a}^m_*$ and annotations $\textbf{a}^m_k$ as follows.
\begin{align}
\label{eqn:a_star_discrete}
\textbf{a}^m_* &= \Theta^T\textbf{x}_m + \bm{\epsilon}_m \\
\label{eqn:ann_discrete}
\textbf{a}^m_k &= F_k\textbf{a}^m_* + \bm{\eta}_k 
\end{align}
where, $\textbf{x}_m \in \rm I\!R^P$; $\Theta \in  \rm I\!R^{P \times D}$; $\bm{\epsilon}_m \sim N(\textbf{0}, \sigma^2I)$; $\sigma^2 \in \rm I\!R$. The annotator noise $\bm{\eta}_k$ is defined as $\bm{\eta}_k \sim N(\textbf{0}, \tau_k^2I)$;  $\tau_k^2 \in \rm I\!R$. $F_k \in  \rm I\!R^{D \times D}$  is the annotator specific weight matrix. Each annotation dimension value $a^{m,d}_k$ for annotator $k$ is defined as a weighted average of the ground truth vector $\textbf{a}^m_*$ with weights given by the vector $F_k(d,:)$.

\subsubsection{Parameter Estimation}

The model parameters $\Phi = \{F_k, \Theta, \sigma^2, \tau_k^2\}$ are estimated using Maximum Likelihood Estimation (MLE) in which they are chosen to be the values that maximize the likelihood function $\mathcal{L}$. 
\begin{align}
\log \mathcal{L} &= \sum_{m=1}^M \log p(\textbf{a}^m_1\dots\textbf{a}^m_K; \Phi) \nonumber \\
&= \sum_{m=1}^M \log \int_{\textbf{a}^m_*} p(\textbf{a}^m_1\dots\textbf{a}^m_K | \textbf{a}^m_*; F_k, \tau_k^2) p(\textbf{a}^m_*; \Theta,\sigma^2)  \, d\textbf{a}^m_* 
\label{eqn:model_likelihood}
\end{align}
Optimizing Equation \ref{eqn:model_likelihood} directly is intractable because of the presence of the integral within the log term, hence we use the EM algorithm. Note that the model we propose assumes that only some random subset of all available annotators provide annotations on a given data instance, as shown in Figure \ref{fig:modelc}. However, for ease of exposition, we overload the variable $K$ and use it here to indicate the number of annotators that attempt to judge the given data instance $m$. 
\subsubsection{EM algorithm}
\label{subsec:disc_em}
The Expectation Maximization (EM) algorithm to estimate the model parameters is shown below. It is an iterative algorithm in which the E and M-steps are executed repeatedly until an exit condition is encountered. Complete derivation of the model can be found in Appendix \ref{appendix_global}. 

\textbf{Initialization} We initialize by assigning the expected values and covariance matrices for the $m$ ground truth vectors $\textbf{a}^m_*$ to their sample estimates (i.e. sample mean and sample covariance) from the corresponding annotations. 
We then estimate the parameters as described in the maximization step using these estimates. 

\textbf{E-step} In this step we take expectation of the log likelihood function with respect to $p(\textbf{a}^m_* | \textbf{a}^m_1\dots\textbf{a}^m_K)$ and the resulting objective is maximized with respect to the model parameters in the M-step. Equations to compute the expected value and covariance matrices for the latent variable $\textbf{a}^m_*$ in the E-step are listed below.
\begin{gather}
\E_{\textbf{a}^m_* | \textbf{a}^m_1\dots\textbf{a}^m_K}[\textbf{a}^m_*] = \Theta^T \textbf{x}_m +  \Sigma_{\textbf{a}^m_*,\textbf{a}^m_1\dots\textbf{a}^m_K} \Sigma^{-1}_{\textbf{a}^m_1\dots\textbf{a}^m_K,\textbf{a}^m_1\dots\textbf{a}^m_K} \nonumber \\(\textbf{a}^m - \boldsymbol{\mu}^m) \nonumber \\
\Sigma_{\textbf{a}^m_* | \textbf{a}^m_1\dots\textbf{a}^m_K} [\textbf{a}^m_*] = \Sigma_{\textbf{a}^m_*,\textbf{a}^m_*} - \Sigma_{\textbf{a}^m_*,\textbf{a}^m_1\dots\textbf{a}^m_K}\Sigma^{-1}_{\textbf{a}^m_1\dots\textbf{a}^m_K,\textbf{a}^m_1\dots\textbf{a}^m_K} \nonumber \\ \Sigma_{\textbf{a}^m_1\dots\textbf{a}^m_K,\textbf{a}^m_*} \nonumber
\end{gather}

The $\Sigma$ terms are covariance matrices between the subscripted random variables. $\textbf{a}^m$ and $\boldsymbol{\mu}^m$ are $DK$ dimensional vectors obtained by concatenating the $K$ annotation vectors $\textbf{a}^m_1,\dots\textbf{a}^m_K$ and their corresponding expected values.

\textbf{M-step} In this step, we compute current estimates for the parameters as follows. The expectations shown below are over the conditional distribution ${\textbf{a}^m_*|\textbf{a}^m_1\dots\textbf{a}^m_K}$.

\begin{gather}
\Theta = (\textrm{X}^T\textrm{X})^{-1}(\textrm{X}^T\E[\textrm{a}^m_*]) \nonumber \\
F_k = \bigg( \sum_{m=1}^{M_k} \textbf{a}^m_K \E[\left(\textbf{a}^m_*\right)^T] \bigg) \bigg( \sum_{m=1}^{M_k}  \E[\textbf{a}^m_*\left(\textbf{a}^m_*\right)^T]\bigg)^{-1} \nonumber \\
\sigma^2 = \frac{1}{md} \sum_{m=1}^{M} \bigg( \E[(\textbf{a}^m_*)^T\textbf{a}^m_*] -   2tr\left(\Theta'^T\textbf{x}_m\E[(\textbf{a}^m_*)^T]\right) \nonumber \\ + tr(\textbf{x}^T_m\Theta'\Theta'^T\textbf{x}_m) \bigg) \nonumber 
\end{gather}
\begin{gather}
\tau_k^2 = \frac{1}{{m_k}d} \sum_{m=1}^{M_k} \bigg( (\textbf{a}^m_K)^T\textbf{a}^m_K - 2tr\big(F_k'^T \textbf{a}^m_K\E[(\textbf{a}^m_*)^T]\big) \nonumber \\ + tr\big(F_k'^TF_k'\E[\textbf{a}^m_*(\textbf{a}^m_*)^T]\big)  \bigg) \nonumber
\end{gather}

Note the similarity of the update equation for $\Theta$ with the familiar normal equations. We are using the soft estimate of $\textbf{a}^m_*$ to find the expression for $\Theta$ in each iteration. Here, $\textrm{X}$ is the feature matrix for all data instances; it includes individual feature vectors $x_m$ in its rows. $\Theta'$ and $F_k'$ are parameters from the previous iteration. 

\textbf{Termination} We run the algorithm until convergence, and stop model training when the change in log-likelihood falls below a threshold of $0.001 \%$. 

\subsection{Time series annotation model}
\label{subsec:cont_model}

In this setting, the ground truth and the annotations are matrices with $T$ rows (time) and $D$ columns (annotation dimensions). The ground truth matrix $\textbf{a}^m_*$ is defined as follows. 
\begin{align}
\label{eqn:contastar}
\text{vec}(\textbf{a}^m_*) &= \text{vec}(\textrm{X}_m\Theta) + \bm{\epsilon}_m
\end{align}

where $\textbf{a}^m_* \in \rm I\!R^{T \times D}$, $\textrm{X}_m \in \rm I\!R^{T \times P}$ and $\Theta \in \rm I\!R^{P \times D}$; $T$ represents the time dimension and is the length of the time series. $\textrm{X}_m$ is the feature matrix where each row corresponds to features extracted from the data instance for one particular time stamp. $\text{vec}(.)$ is the vectorization operation which flattens the input matrix in column first order to a vector. $\bm{\epsilon}_m \sim \mathcal{N}(\bm{0},\sigma^2I) \in \rm I\!R^{TD}$  is the additive noise vector with $\sigma \in \rm I\!R$.

In \cite{Gupta2015}, the authors propose a linear model where the annotation function $g(\textbf{a}^m_*;F_k)$ is a causal linear time invariant (LTI) filter of fixed width. The advantage of using an LTI filter is that it can capture scaling and time-delay biases introduced by the annotators. 

The filter width $W$ is chosen such that $W\ll T$, where $T$ is the number of time stamps for which we have the annotations. The annotation function for dimension $d'$ can be viewed as the left multiplication of a filter matrix $B^{d'}_k \in \rm I\!R^{T \times T}$ as shown in Equation \ref{eqn:filter_matrix}.

\begin{equation}
B^{d'}_k =
\begin{bmatrix}
	b^{d'}_1 & 0 & 0 & 0 & 0 & \dots & 0 \\
    b^{d'}_2 & b^{d'}_1 & 0 & 0 & 0 & \dots & 0 \\
    b^{d'}_3 & b^{d'}_2 & b^{d'}_1 & 0 & 0 & \dots & 0 \\
    \vdots & \vdots & \vdots & \ddots & \vdots \\    
    0 & b^{d'}_W & \dots & b^{d'}_1 & 0 & \dots & 0 \\
    \vdots & \vdots & \vdots & \ddots & \vdots \\
    0 & 0 & 0 & 0 & b^{d'}_W & \dots & b^{d'}_1 \\
\end{bmatrix}
\label{eqn:filter_matrix}
\end{equation}

We extend this model in our work to combine information from all of the annotation dimensions. Specifically, the ground truth is left multiplied by $D$ horizontally concatenated filter matrices, each $\in \rm I\!R^{T \times T}$ corresponding to a different dimension as shown below.

\begin{align}
\textbf{a}^{m,d}_k &= F^d_k\text{vec}(\textbf{a}^m_*) + \bm{\eta}_k \\
\text{where, } \nonumber \\ F^d_k &= [B^{d,1}_k, B^{d,2}_k, \dots, B^{d,D}_k] 
\label{eqn:F_k_defn}
\end{align} 

\noindent $F^d_k \in \rm I\!R^{T \times TD}$ with $WD$ unique parameters. $\bm{\eta}_k \sim \mathcal{N}(\bm{0}, \tau^2_kI)\in \rm I\!R^{T}$ with $\tau^2_k \in \rm I\!R$.

\subsubsection{Parameter Estimation} Estimating the model parameters similar to the global model requires computing the expectations over a vector of size $TD$. Since $T$ is the number of time stamps in the task and can be arbitrarily long, this may not be feasible in all tasks. For example, in the movie emotions corpus \cite{malandrakis2011supervised}, annotations are computed at a rate of 25 frames per second with each file of duration $\sim$30 minutes or of $\sim$45k annotation frames. To avoid this we use a variant of EM named \textit{Hard EM} in which instead of taking expectations over the entire conditional distribution of $\textbf{a}^m_*$ we find its mode. This variant has been shown to be comparable in performance to the classic EM (\textit{Soft EM}) despite being significantly faster and simple \cite{spitkovsky2010viterbi}. This approach is similar to the parameter estimation strategy devised by \cite{Gupta2015} in their time series annotation model.

The likelihood function is similar to the global model in Equation \ref{eqn:model_likelihood} as shown below.
\begin{align}
\log \mathcal{L} &= \sum_{m=1}^M \log \int_{\textbf{a}^m_*} p(\textbf{a}^m_1\dots\textbf{a}^m_K | \textbf{a}^m_*; F_k, \tau_k^2) p(\textbf{a}^m_*;\Theta, \sigma^2)  \, d\textbf{a}^m_* \nonumber
\end{align}

However the integral here is with respect to the flattened vector $\text{vec}(\textbf{a}^m_*)$.

\subsubsection{EM algorithm} 
\label{subsec:cont_em}
The EM algorithm for the time series annotation model is listed below. Complete derivations can be found in Appendix \ref{appendix_timeseries}.

\textbf{Initialization} Unlike the global annotation model, we initialize $\textbf{a}^m_*$ randomly since we observed better performance when compared to initializing it with the annotation means. Given this $\textbf{a}^m_*$, the model parameters are estimated as described in the maximization step below.

\textbf{E-step} In this step we assign $\textbf{a}^m_*$ to the mode of the conditional distribution $q(\textbf{a}^m_*) = p(\textbf{a}^m_*|\textbf{a}^m_1, \dots, \textbf{a}^m_K)$. Since this distribution is normal (see appendix \ref{appendix_global}) finding the mode is equivalent to minimizing the following expression. 

\begin{gather}
\textbf{a}^m_* = \argmin_{\textbf{a}^m_*} \sum_k \sum_d ||\textbf{a}^{m,d}_k - F^d_k \text{vec}(\textbf{a}^m_*)||^2_2 + \nonumber \\ ||\text{vec}(\textbf{a}^m_*) - \text{vec}(\textrm{X}_m\Theta)||^2_2 \nonumber
\end{gather}

\label{cont:mstep}
\textbf{M-step} Given the estimate for $\textbf{a}^m_*$ from the E-step, we substitute it in the likelihood function and maximize with respect to the parameters in the M-step. The estimates for the different parameters are shown below.

\begin{gather}
\Theta = \bigg(\sum_{m=1}^M \textrm{X}_m^T\textrm{X}_m\bigg)^{-1}\bigg(\sum_{m=1}^M\textrm{X}_m^T\textbf{a}^m_*\bigg) \nonumber \\
f^d_k = \bigg(\sum_{m=1}^{M_k} A^TA\bigg)^{-1}\bigg(\sum_{m=1}^{M_k} A^T\textbf{a}^{m,d}_k\bigg) \nonumber \\
\sigma^2 = \frac{1}{MTD}\sum_{m=1}^{M} ||\text{vec}(\textbf{a}^m_K) - \text{vec}(\textrm{X}_m\Theta)||^2_2 \nonumber \\
\tau^2_k = \frac{1}{M_kTD}\sum_{m=1}^{M_k} \sum_{d} ||\textbf{a}^{m,d}_k - F^d_k\text{vec}(\textbf{a}^m_*)||^2_2 \nonumber
\end{gather}

\noindent $M_k$ is the number of files annotated by user $k$; $A$ is a matrix obtained by reshaping $\text{vec}(\textbf{a}^m_*)$ as described in subsection \ref{subsec:cont_mstep}.

\textbf{Termination} We run the algorithm until convergence, and stop model training when the change in log-likelihood falls below a threshold of $0.5 \%$.

\section{Experiments \& Results}
\label{sec:exp}

We evaluate the models described above on three different types of data: synthetic data, an artificial task with human annotations, and finally with real data. We describe these below. We compare our joint models with their \textit{independent} counterparts as baselines, in which each annotation dimension is modeled separately. This allows us to highlight the benefits of moving to a multi-dimensional annotation fusion scheme with everything else kept constant. Update equations for the independent model can be obtained by running the models described above for each dimension separately with $D=1$. Note that the independent model is similar in the global setting to the regression model proposed in \cite{raykar2010learning} (with ground truth scaled by the singleton $f^d_k$). In the time series setting it is identical to the model proposed by \cite{Gupta2015}.

The models are evaluated by comparing the estimated $\textbf{a}^m_*$ with the actual ground truth. We report model performance using two metrics: the Concordance correlation coefficient ($\rho_c$) \cite{lawrence1989concordance} and the Pearson's correlation coefficient ($\rho$). $\rho_c$ measures any departures from the \textit{concordance line} (line passing through the origin at $45\degree$ angle). Hence it is sensitive to rotations or rescaling in the predicted ground truth. Given two samples $x$ and $y$, the sample concordance coefficient $\hat{\rho_c}$ is defined as shown below.

\begin{equation}
\hat{\rho_c} = \frac{2s_{xy}}{s^2_x+s^2_y+(\bar{x} - \bar{y})^2} \nonumber
\end{equation}

\noindent We also report results in Pearson's correlation to highlight the accuracy of the models in the presence of rotations. 

As noted before, the models proposed in this paper are closely related to the Factor Analysis model, which is vulnerable to issues of unidentifiability \cite{fabrigar1999evaluating}, due to the matrix factorization. Different types of unidentifiability have been studied in literature, such as factor rotation, scaling and label switching. 
In our experiments, we handle label switching through manual judgment (by reassigning the estimated ground truth between dimensions if necessary) as is common in psychology \cite{kaiser1958varimax}, but defer the task of choosing an appropriate prior on the rotation matrix $F_k$ to address other unidentifiabilities for future work. 

We report aggregate test set results using $C$-fold cross validation. To address overfitting, within each fold, we evaluate the parameters obtained after each iteration of the EM algorithm by estimating the ground truth on a disjoint validation set, and pick those with the highest performance in concordance correlation $\rho_c$ as the parameter estimates of the model. We then estimate the performance of this parameter set in predicting the ground truth from a separate held out test set for that fold. Finally, we also report statistically significant differences between the joint and independent models at $5 \%$ false-positive rate ($\alpha = 0.05$) in all our experiments. 

\subsection{Global annotation model}
The global annotation model uses the EM algorithm described in Section \ref{subsec:disc_em} to estimate the ground truth for discrete annotations. We evaluate the model in three different settings described below. Statistical significance tests were run by computing bootstrap confidence intervals \cite{efron1994introduction} on the differences in model performances across the $C$-folds. To establish the statistical significance, we ran the joint and independent models to obtain $C$ test set model predictions from $C$ folds. Given these, we ran $1000$ bootstrap iterations in which the test set predictions were sampled with replacement, from which $\rho$ and $\rho_c$ were estimated for each dimension. We conclude significance if the evaluation metric being examined was higher in at least $95\%$ of the bootstrap runs. 

\begin{figure}
    \centering
    \begin{subfigure}[t]{0.45\linewidth}
        \centering 
        \includegraphics[scale=0.7,trim={1.5cm 0 1cm 0.5cm}]{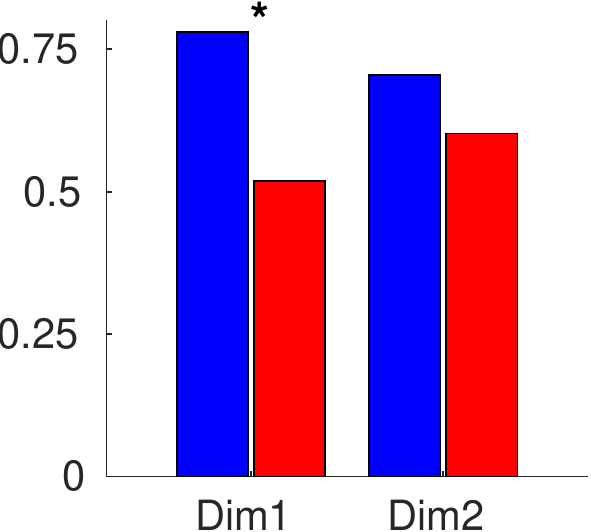}
        \caption{Concordance ($\rho_c$)}
    \end{subfigure}%
    ~
    \begin{subfigure}[t]{0.45\linewidth}
        \centering 
        \includegraphics[scale=0.7,trim={0.5cm 0 1cm 0.5cm}]{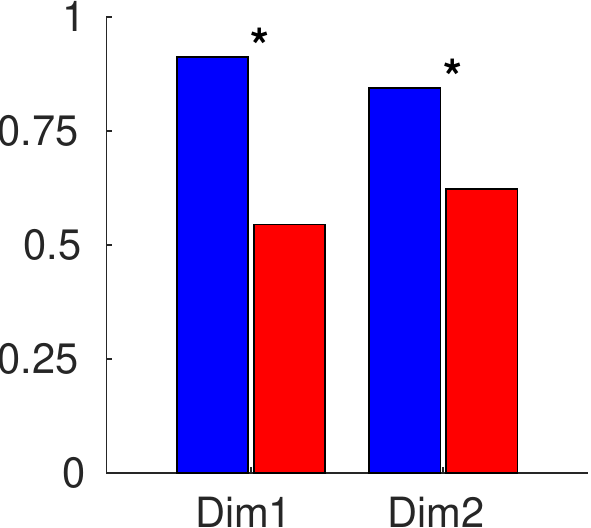}
        \caption{Pearson ($\rho$)}
    \end{subfigure}
    \par\medskip
    \begin{subfigure}[\textwidth]{0.5\textwidth}
        \includegraphics[scale=0.3,trim=11.5cm 4.5cm 1cm 4cm,clip=true]{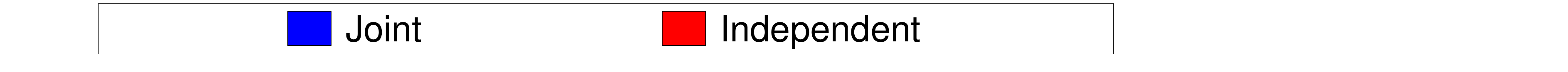}
    \end{subfigure}
    \caption{Performance of global annotation model on synthetic dataset; \textit{*-statistically significant}}
    \label{fig:disc_res_syn}
\end{figure}

\subsubsection{Synthetic data}
\label{subsec:disc_syn}
We created synthetic data according to the model described in Section \ref{subsec:disc_model} with random features $\textrm{X} \in \rm I\!R^{500}$ for 100 data instances each with 2 dimensions of annotations (i.e. $D$=2). 10 artificial annotators, each with unique random $F_k$ matrices were used to produce annotations for all the data instances. Elements of the feature matrices were sampled from the standard normal distribution, while the elements of $F_k$ matrices were sampled from $\mathcal{U}(0,1)$. Elements of ground truth $\textbf{a}^m_*$ were sampled from $\mathcal{U}(-1,1)$ and $\theta$ was estimated from $\textbf{a}^m_*$ and $\textrm{X}$. Since its off diagonal elements are non-zero, our choice of $F_k$ represents tasks in which the annotation dimensions are related to each other.

Figure \ref{fig:disc_res_syn} shows the performance of joint and independent models in predicting the ground truth $\textbf{a}^m_*$. For both dimensions, the proposed joint model predicts the $\textbf{a}^m_*$ with considerably higher accuracy as shown by the higher correlations, highlighting the advantages of modeling the annotation dimensions jointly when they are expected to be related to each other. 

\begin{figure}
    \centering
    \begin{subfigure}[t]{0.45\linewidth}
        \centering 
        \includegraphics[scale=0.7,trim={1.5cm 0 1cm 0.5cm}]{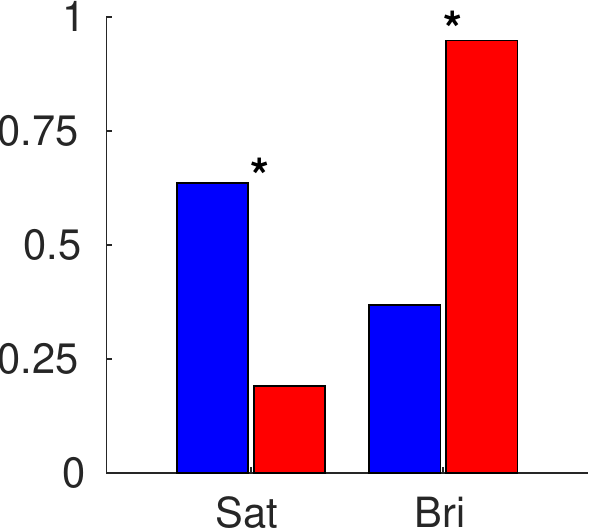}
        \caption{Concordance ($\rho_c$)}
    \end{subfigure}%
    ~
    \begin{subfigure}[t]{0.45\linewidth}
        \centering 
        \includegraphics[scale=0.7,trim={0.5cm 0 1cm 0.5cm}]{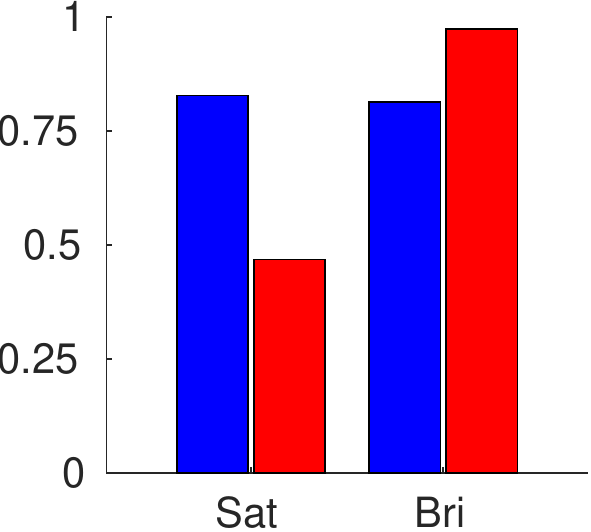}
        \caption{Pearson ($\rho$)}
    \end{subfigure}
    \par\medskip
    \begin{subfigure}[\textwidth]{0.5\textwidth}
        \includegraphics[scale=0.3,trim=11.5cm 4.5cm 1cm 4cm,clip=true]{bar_legend}
    \end{subfigure}
    \caption{Performance of global annotation model on artificial dataset; \textit{Sat-Saturation, Bri-Brightness; *-statistically significant}}
    \label{fig:disc_res_art}
\end{figure}

\subsubsection{Artificial data}

Since crowdsourcing experiments typically involve collecting subjective annotations, they seldom have well defined ground truth. As a result, most annotation models are evaluated on expert annotations collected by specially trained users. 
For example, while collecting annotations on medical data, labels estimated by fusing annotations from naive users may be evaluated against those provided by experts such as doctors. 
However, this poses a circular problem since the expert annotations themselves may be subjective and combining them to estimate the ground truth is not straightforward. To address this, we created an artificial task with controlled ground truth on which we collect annotations from multiple annotators and evaluate the fused annotation values with the known ground truth values, similar to \cite{booth2018novel}. In our task, the annotators were asked to provide their best estimates on perceived saturation and brightness values for monochromatic images. The relationship between perceived saturation and brightness is well known as the Helmholtz\textemdash Kohlrausch effect \cite{corney2009brightness}, according to which, increasing the saturation of an image leads to an increase in the perceived brightness, even if the actual brightness was constant. 

In our experiments, we collected annotations on images from two regimes: one with fixed saturation and varying brightness, and vice versa. This approach was chosen since it would allow us to evaluate the impact of change in either brightness or saturation while the other was held constant. The color of the images were chosen randomly (and independent of the image's saturation and brightness) between green and blue. Annotations were collected on Mturk and the annotators were asked to familiarize themselves with saturation and brightness using an online interactive tool before providing their ratings. In both experiments, a reference image with fixed brightness and saturation was inserted after every ten annotation images to prevent any bias in the annotators. The reference images were hidden from the annotators and appeared as regular annotation images. For parameter estimation, RGB values were chosen as the features for each image.

We used the joint model to estimate the ground truth for the two regimes separately since we expect the relationship between saturation and brightness to be dissimilar in the two cases. From each experiment, predicted values of the underlying dimension being varied was compared with the actual $\textbf{a}^m_*$ values. For example, in the experiment with varying saturation and fixed brightness, the joint model was run on full annotations, but only the estimated values of saturation were compared with ground truth saturation. For the independent model, we use annotation values of the underlying dimension being varied from each regime, and compare the estimated values with ground truth.

Figure \ref{fig:disc_res_art} shows the performance of the joint and independent models for this experiment. The joint model leads to better estimates of saturation when compared to the independent model by making use of the annotations on brightness. This agrees with the Helmholtz\textemdash Kohlrausch phenomenon described above, since the annotators can perceive the changing saturation as a change in brightness, leading to correlated annotations for the two dimensions. On the other hand, the independent model leads to better estimates of brightness, which seems to have no effect on perceived saturation annotations. This experiment highlights the benefits of jointly modeling annotations in cases where the annotation dimensions may be correlated or dependent on one another.

\subsubsection{Real data}
\begin{figure}
    \centering
    \begin{subfigure}[t]{0.8\linewidth}
        \centering 
        \includegraphics[scale=0.45,trim={1.5cm 0 1cm 0.5cm}]{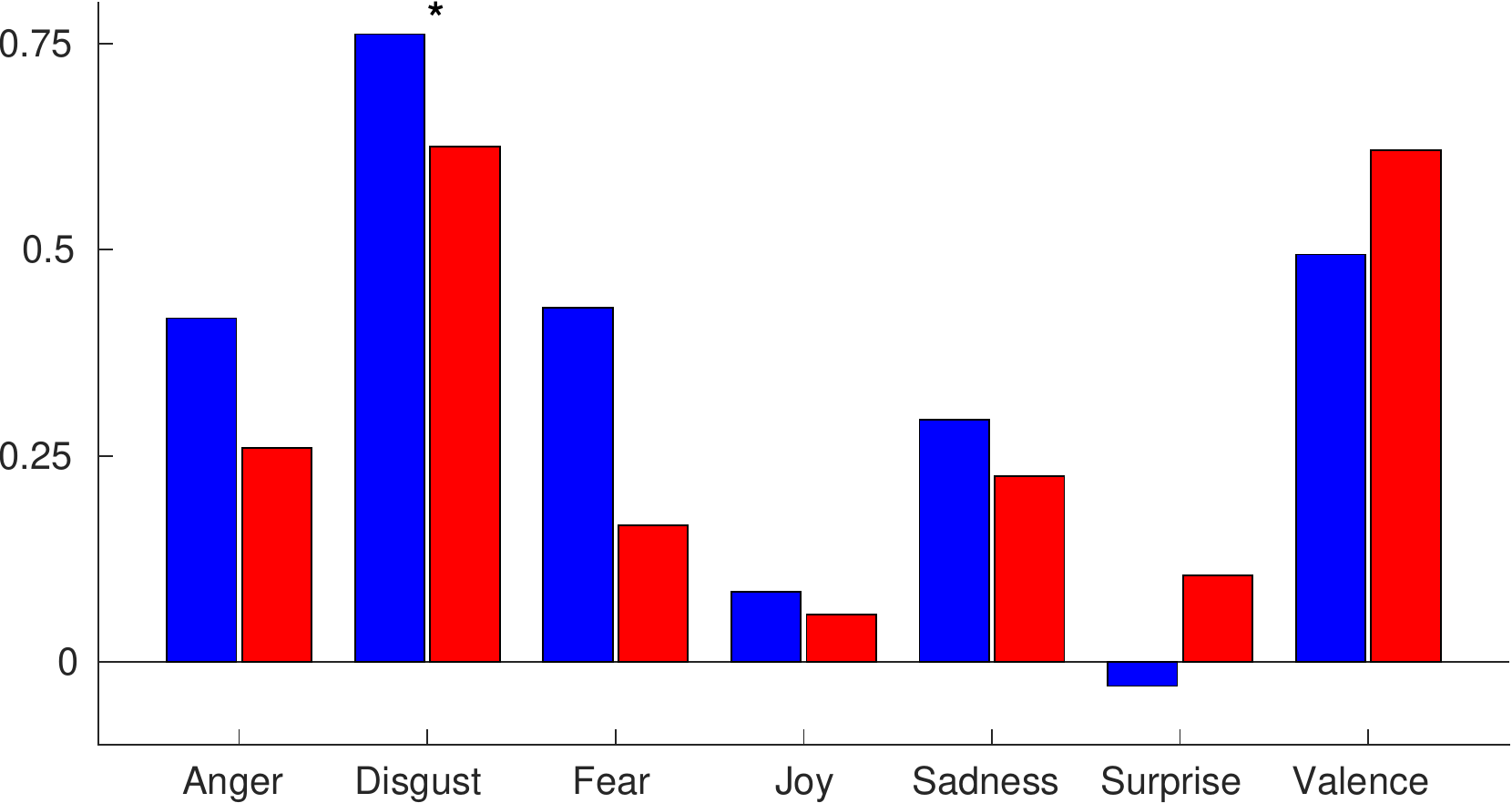}
        \caption{Concordance correlation ($\rho_c$)}
    \end{subfigure}
    \par\medskip
    \begin{subfigure}[t]{0.8\linewidth}
        \centering 
        \includegraphics[scale=0.45,trim={1.5cm 0 1cm 0.5cm}]{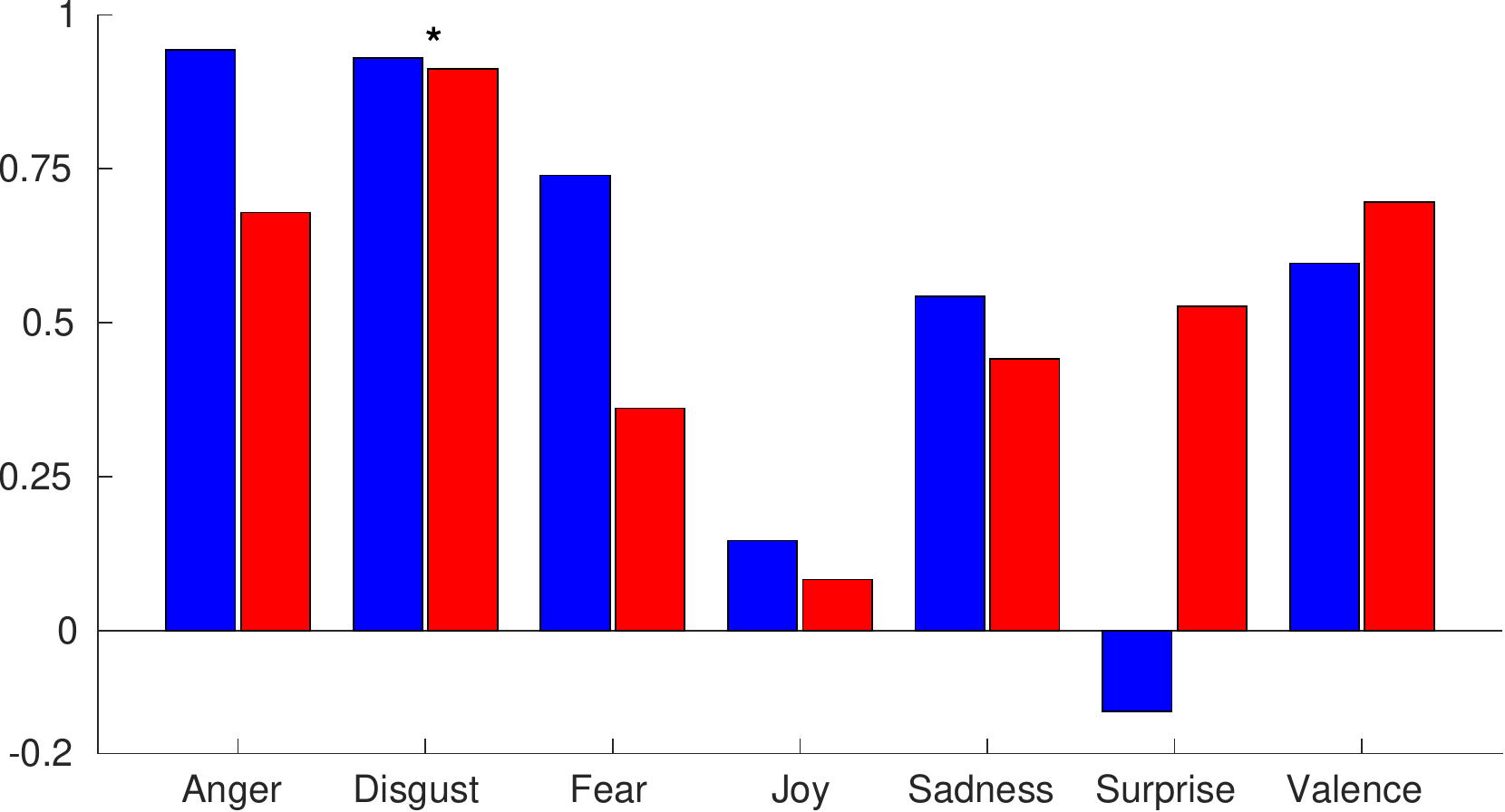}
        \caption{Pearson correlation ($\rho$)}
    \end{subfigure}
    \par\medskip
    \begin{subfigure}[\textwidth]{0.5\textwidth}
        \includegraphics[scale=0.3,trim=11.5cm 4.5cm 1cm 4cm,clip=true]{bar_legend}
    \end{subfigure}
    \caption{Performance of global annotation model on the text emotions dataset; \textit{*-statistically significant}}
    \label{fig:disc_res_real}
\end{figure}

Our final experiment for the global model was on the task of annotating news headlines in which the annotators provide numeric ratings for various emotions. This dataset was first described in the 2007 SemEval task on affective text \cite{Strapparava:2007:STA:1621474.1621487}. Numeric ratings from the original task were labeled by trained annotators and we treat these as expert annotations. We use Mturk annotations from \cite{snow2008cheap} as the actual input to our model. Sentence level annotations are provided on seven emotions ($D$=7): \textit{anger}, \textit{disgust}, \textit{fear}, \textit{joy}, \textit{sadness}, \textit{surprise} and \textit{valence} (positive/negative polarity). We use sentence level embeddings computed using the pre-trained sentence embedding model sent2vec\footnote{https://github.com/epfml/sent2vec}  \cite{DBLP:journals/corr/PagliardiniGJ17} as feature vectors $x$ for the model.

Figure \ref{fig:disc_res_real} shows the performance of the joint and independent models on this task. The joint model shows better performance in predicting the reference emotion labels for \textit{anger}, \textit{disgust}, \textit{fear}, \textit{joy} and \textit{sadness}, but performs worse than the independent model in predicting \textit{surprise} and \textit{valence}. 

\begin{figure*}
\centering
\begin{subfigure}{\linewidth}
\centering
\includegraphics[scale=0.9,trim={1cm 0 0 0cm}]{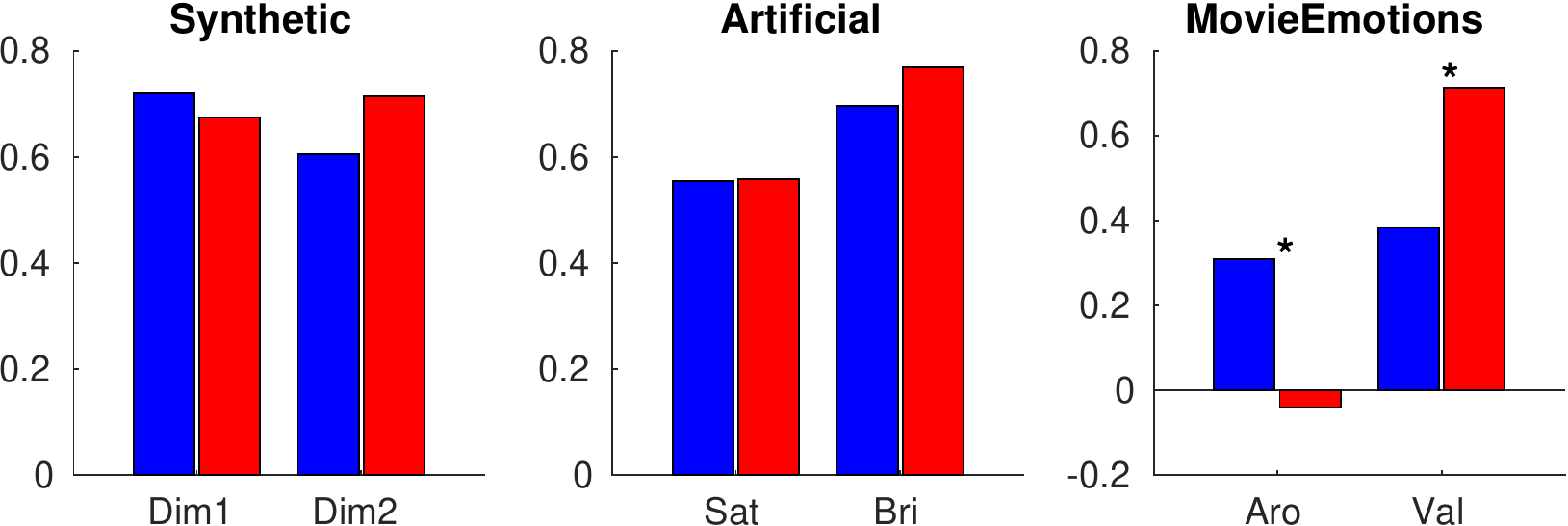}
\caption{Concordance correlation coefficients}
\end{subfigure}
\par\medskip
\begin{subfigure}{\linewidth}
\centering
\includegraphics[scale=0.9,trim={1cm 0 0 0cm}]{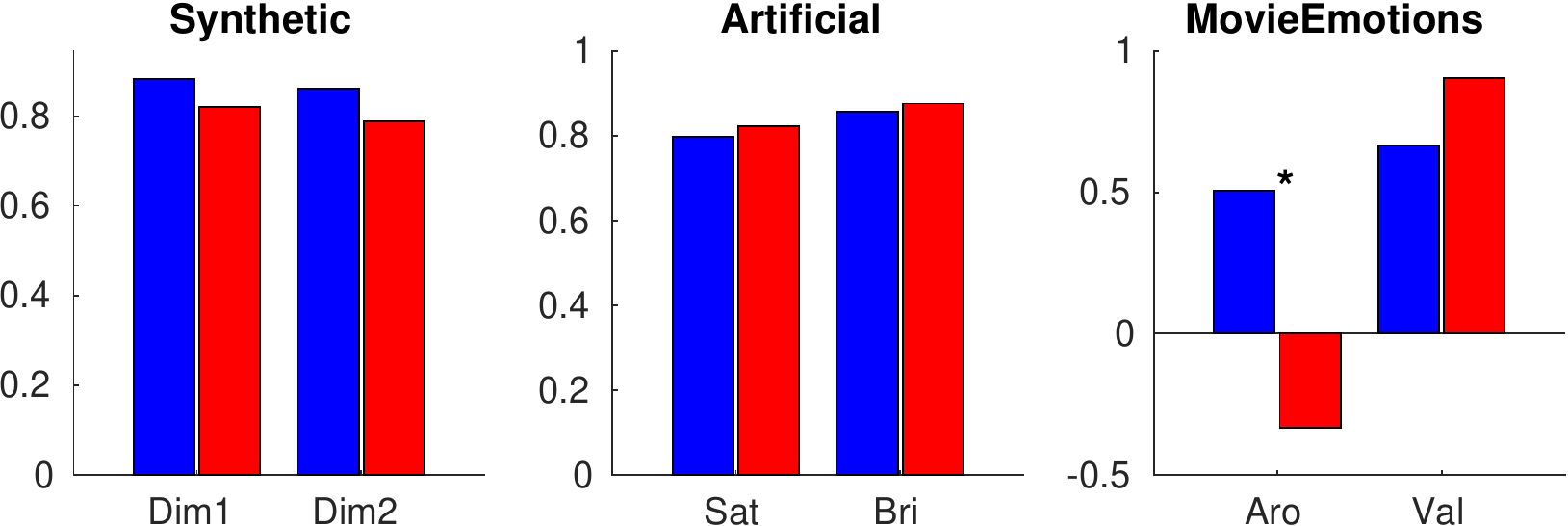}
\caption{Pearson correlation coefficients}
\end{subfigure}
\par\medskip
\begin{subfigure}{\linewidth}
\centering
\includegraphics[scale=0.5,trim=8cm 4.5cm 0cm 4cm,clip=true]{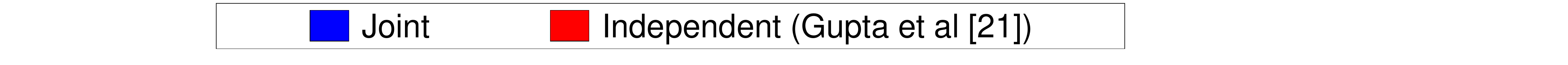}
\end{subfigure}
\caption{Concordance and Pearson correlation coefficients between ground truth/reference and model predictions for the time series annotation model; \textit{*-statistically significant}}
\label{fig:cont_results}
\end{figure*}

\subsection{Time series annotation model}
In this setting, the annotations are collected on data with a temporal dimension, such as time series data, video or audio signals. Similar to the global model, we evaluate this model in 3 settings: synthetic, artificial and on real data. The evaluation metrics $\rho_c$ and $\rho$ are computed over estimated and actual ground truth vectors $\textbf{a}^m_*$ by concatenating the data instances into a single vector. The time series models have the window size $W$ as an additional hyperparameter, which is selected using a validation set. In each fold of the dataset, we train model parameters for different window sizes from the set $\{5,10,20,50\}$, and pick $W$ and related parameters with the highest concordance correlation $\rho_c$ on the validation set. These are then evaluated on a disjoint test set, and we repeat the process for each fold. In each experiment, the parameters were initialized randomly, and the process was repeated 20 times at different random initializations, selecting the best starting point using the validation set. 
To identify significant differences, we compute the test set performance of the two models for each fold, and run the paired t-test between the $C$ sized samples of $\rho$ and $\rho_c$ corresponding to the joint and independent models. We do not bootstrap confidence intervals due to smaller test set sizes.

\subsubsection{Synthetic data}
\label{subsec:cont_syn}
The synthetic dataset was created using the model described in Section \ref{subsec:cont_model}. Elements of the feature matrix were sampled from the standard normal distribution while elements of $F_k$ and ground truth were sampled from $\mathcal{U}(0,1)$. In this setting each data instance includes $T$ feature vectors, one for each time stamp. The time dependent feature matrices were created using a random walk model without drift but with \textit{lag} to mimic a real world task. In other words, while creating the $P$ dimensional time series, the features vectors were held fixed for a time period arbitrarily chosen to be between 2 to 4 time stamps. This was done because in most tasks the underlying dimension (such as emotion) is expected to remain constant at least for a few seconds. In addition, the transition between changes in the feature vectors were linear and not abrupt. In our experiments, we chose $P=500$, $T=350$, $D=2$, $M=18$ and the number of annotators $K=6$. 

Figure \ref{fig:cont_results} shows the aggregate results across $C$-folds ($C=5$) for the joint and independent models in the 3 settings. In the synthetic dataset, the joint model achieves higher values for Pearson's correlation $\rho$ for both the dimensions and higher value for $\rho_c$ for dimension 1. For dimension 2 however, the independent model achieves better $\rho_c$. 

\subsubsection{Artificial data}
We collected annotations on videos with the artificial task of identifying saturation and brightness, described in the previous section. The videos consisted of monochromatic images with the underlying saturation and brightness varied independent of each other. The dimensions were created using a random walk model with lag as described in Section \ref{subsec:cont_syn}. The annotations were collected in house using an annotation system developed using the Robot Operating System \cite{Quigley09}. 10 graduate students gave their ratings on the two dimensions. Each dimension was annotated independently using a mouse controlled slider. For parameter estimation, the feature vectors for each time stamp were RGB values.

As seen in Figure \ref{fig:cont_results}, both models achieve similar performance in predicting the ground truth for saturation and brightness in terms of $\rho$, as well as in predicting saturation in terms of $\rho_c$. The independent model achieves slightly better performance in predicting brightness in terms of concordance correlation (though not statistically significant); however, their performance in terms of $\rho$ suggests that the joint model output differs only in terms of a linear scaling. 
The joint model appears to be at par with the independent model for the most part, suggesting that the transformation matrix $F_k$ connecting the two dimensions for each annotator, is unable to accurately capture the dependencies between the dimensions, likely due to the fact that, unlike the global annotation model, the underlying brightness and saturation were varied simultaneously and independent of each other (leading to non-linear dependencies between them), and that we limit $F_k$ to only capture linear relationships.

\subsubsection{Real data}
We finally evaluate our model on a real world task with time series annotations. We chose the task of predicting the affective dimensions of valence and arousal from movie clips, first described in \cite{malandrakis2011supervised}. The associated corpus includes time series annotations of valence and arousal on contiguous 30 minute video segments from 12 Academy Award winning movies. This task was chosen because the data set includes both expert annotations as well as annotations from naive users. We treat the expert annotations as reference and evaluate the estimated dimensions against them; however, we note that the expert labels were provided by just one annotator, which may itself be noisy.

For each movie clip, 6 annotators provide annotations on their \textit{perceived} valence and arousal using the Feeltrace \cite{cowie2000feeltrace} annotation tool. The features used in our parameter estimation include combined audio and video features extracted separately. The audio features were estimated using the emotion recognition baseline features from Opensmile \cite{Eyben:2010:OMV:1873951.1874246} at 25 fps (same frame rate as the video clips) and aggregated at a window size of 5 seconds using the following statistical functionals: mean, max, min, std, range, kurtosis, skewness and inter-quartile range. The video features were extracted using OpenCV \cite{opencv_library} and included frame level luminance, intensity, Hue-Saturation-Value (HSV) color histograms and optical flow \cite{chen1998multimodal}, which were also aggregated to 5 seconds using simple averaging. The combined features were of size $P=1225$ for each frame. 

Figure \ref{fig:cont_results} shows the performance of the two models in estimating the affective dimensions for the dataset. The joint model seems to considerably outperform the independent model while estimating arousal while the independent models seem to produce better estimates of valence from the annotations. 
The independent model seems to perform poorly in arousal prediction, but the joint model shows a balanced performance, with the joint modeling constraint likely acting as a regularizer.

\begin{figure}
\centering
\begin{subfigure}[b]{0.45\textwidth}
\includegraphics[width=\textwidth]{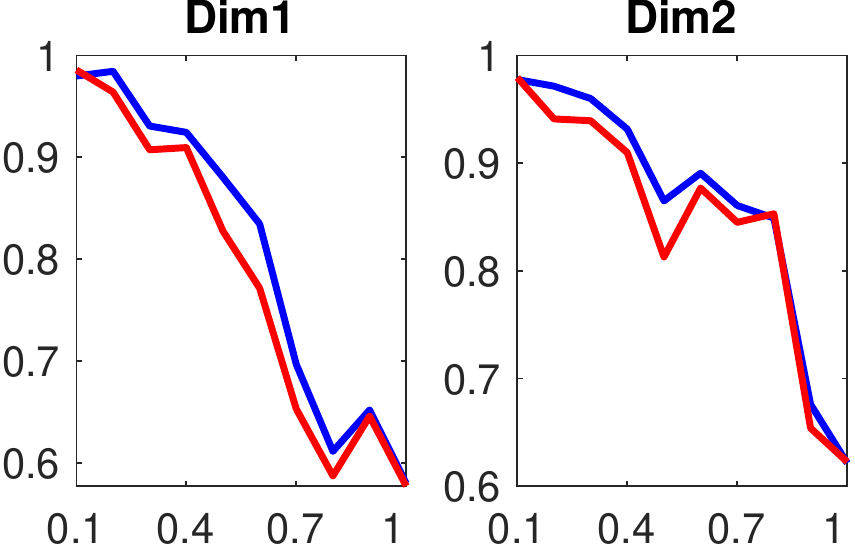}
\caption{Concordance correlation}
\end{subfigure}
\par\medskip
\begin{subfigure}[b]{0.45\textwidth}
\includegraphics[width=\textwidth]{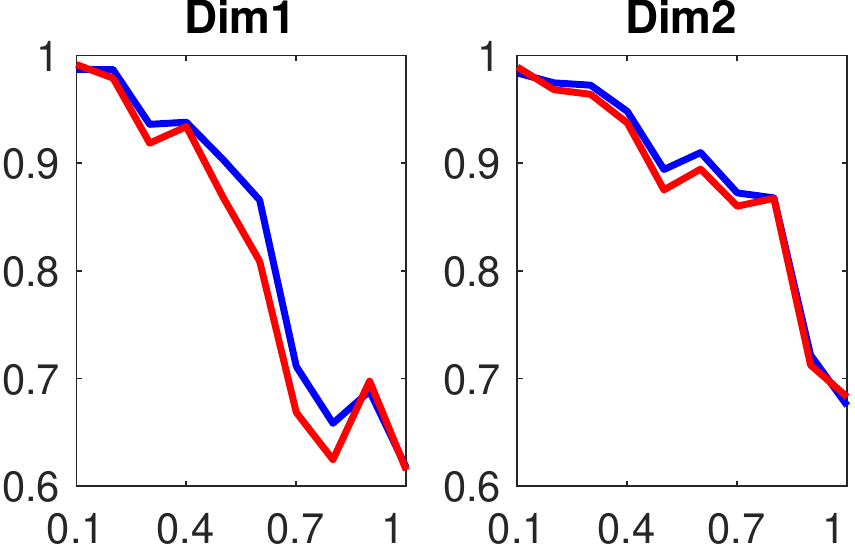}
\caption{Pearson's correlation}
\end{subfigure}
\par\medskip
\begin{subfigure}[\textwidth]{0.5\textwidth}
\includegraphics[scale=0.3,trim=11.5cm 4.5cm 1cm 4cm,clip=true]{bar_legend}
\end{subfigure}
\caption{Effect of varying dependency between annotation dimensions for the synthetic model}
\label{fig:disc_dependency}
\end{figure}

\subsection{Effect of dependency among dimensions}
\label{subsec:varying_fk}

To evaluate the impact of the magnitude of dependency between the annotation dimensions on the performance of the models, we created a set of synthetic annotations for the global model similar to Section \ref{subsec:disc_syn}. We created 10 synthetic datasets, each with constant $F_k$ matrices across all annotators. The principal diagonal elements were fixed to 1 while the off diagonal elements were increased between 0.1 to 1 with a step size of 0.1. Similar to the previous setting, we created 100 annotators, each operating on 10 files. Note that despite the annotators having identical $F_k$ matrices, their annotations on a given file were different because of the noise term $\bm{\eta}_k$ in Equation \ref{eqn:ann_eqn}. 

Figure \ref{fig:disc_dependency} shows the 5-fold cross validated performance of the joint and independent models on this task. As seen in the figure, the joint model consistently outperforms the independent model in both metrics. Both the models start with similar performance when the off diagonal elements are close to zero since this implies no dependency between the annotation dimensions, and the performance of both models continues to degrade as the off diagonal elements increase. However, the joint model is able to make better predictions of the ground truth by making use of the dependency between the dimensions, highlighting the benefits of modeling the annotation dimensions jointly. Visualizations for averaged estimates of the $F_k$ matrices from this experiment can be found in Section \ref{appendix_analysis_fk}.

\section{Conclusion}
\label{sec:conc}

We presented a model to combine multi-dimensional annotations from crowdsourcing platforms such as Mturk. The model assumes the ground truth to be latent and distorted by the annotators. The latent ground truth and the model parameters are estimated using the EM algorithm. EM updates are derived for both global and time series annotation settings. We evaluate the model on synthetic and real data. We also propose an artificial task with controlled ground truth and evaluate the model. 

Weaknesses of the model include vulnerability to unidentifiability issues like most variants of factor analysis \cite{fabrigar1999evaluating}. 
Typical strategies to address this issue involve adapting a suitable prior constraint on the factor matrix. For example, in PCA, the factors are ordered such that they are orthogonal to each other and arranged in decreasing order of variance. 
In our experiments, the model was found to be vulnerable to unidenfiability due to label switching, which was addressed through manual judgements. We defer the task of choosing an appropriate prior constraint on $F_k$ for future work.

Future work includes generalizing the model with Bayesian extensions, in which case the parameters can be estimated using variational inference, in addition to adding model constraints to ensure identifiability of all model parameters. Though we limit our analysis here to linear relationships between the transformation matrix $F_k$ and the ground truth vector $\textbf{a}^m_*$, we note that extending the model to capture non-linear relationships is straightforward. For example, the vector $\textbf{a}^m_*$ in Equation \ref{eqn:ann_eqn} can be replaced by one that includes a non-linear dependence on $\textbf{a}^m_*$.  Providing theoretical bounds to the model performance, specially with respect to the sample complexity may also be possible since we have assumed normal distributions throughout the model.

\section{Acknowledgements}
\label{sec: ack}
The authors would like to thank Zisis Skordilis for all the helpful discussions and feedback.

\bibliographystyle{IEEEtran}
\bibliography{references}

\begin{IEEEbiography}[{\includegraphics[width=1in,height=1.25in,clip,keepaspectratio]{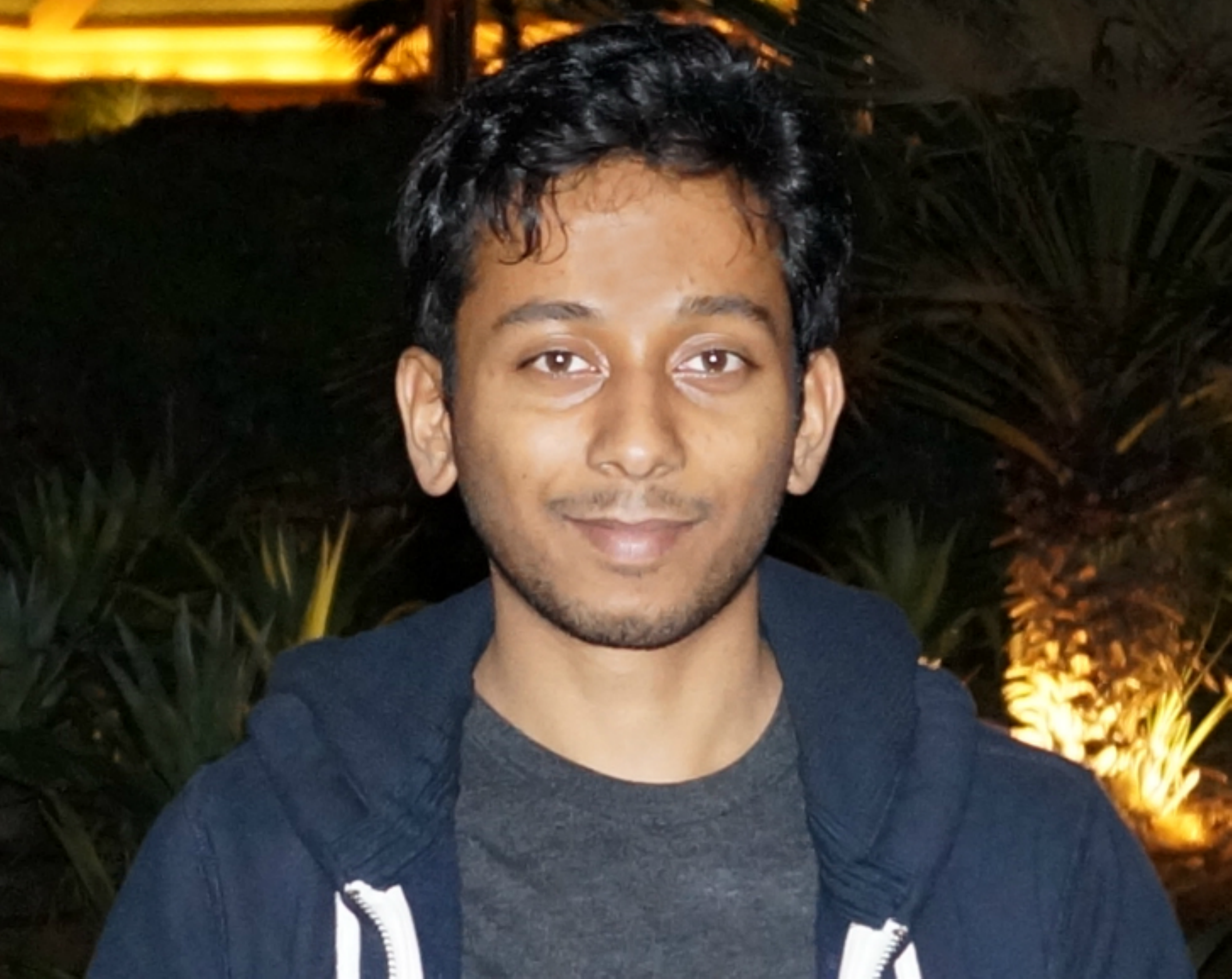}}]{Anil Ramakrishna}
received his B.E. degree from the Visvesvaraya Technological University in 2010, M.S in Computer Science in 2014, M.S in Electrical Engineering in 2019, Ph.D. in Computer Science in 2019, all from the University of Southern California (USC). His dissertation focused on developing computational models for multidimensional annotation fusion, leveraging relationships between annotation dimensions in estimating the ground truth, which is modeled as a latent variable. 
His research interests include sentiment analysis, natural language processing, machine learning and more recently, spoken language understanding. He is a member of the IEEE.
\end{IEEEbiography}

\begin{IEEEbiography}[{\includegraphics[width=1in,height=1.25in,clip,keepaspectratio]{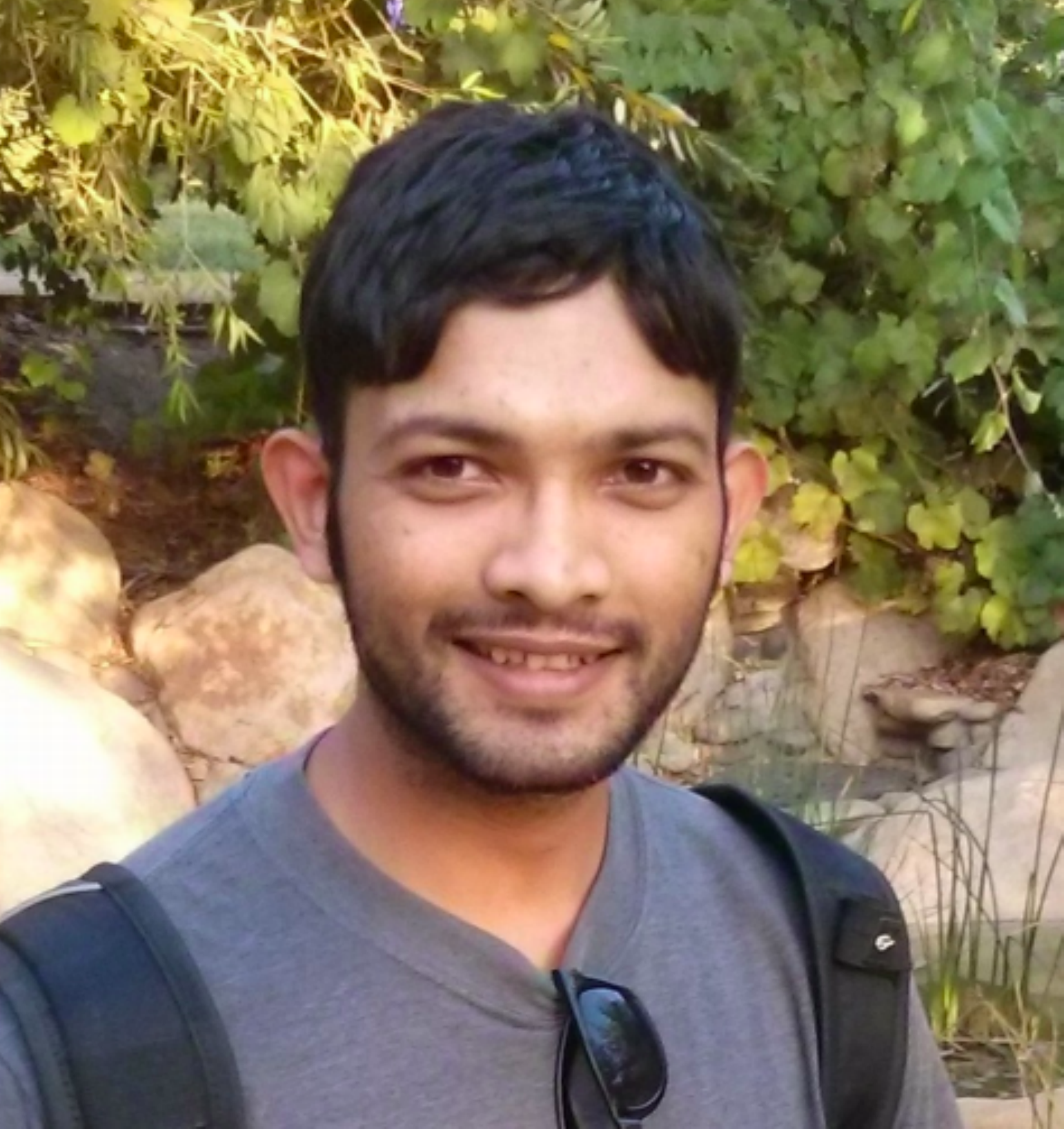}}]{Rahul Gupta} received a B.Tech. degree in Electrical Engineering from Indian Institute of Technology, Kharagpur in 2010 and a Ph.D. degree in Electrical Engineering from University of Southern California (USC), Los Angeles in 2016. His research concerns development of machine learning algorithms with application to human behavioral data. His dissertation work is on the development of computational methods for modeling non-verbal communication in human interaction. He is the recipient of Info-USA exchange scholarship (2009), Provost fellowship (2010-2014) and the Phi Beta Kappa alumni in Southern California scholarship (2015). He was part of the team that won the INTERSPEECH-2013 and INTERSPEECH-2015 Computational Paralinguistics Challenges. He is a member of the IEEE.
\end{IEEEbiography}

\begin{IEEEbiography}[{\includegraphics[width=1in,height=1.25in,clip,keepaspectratio]{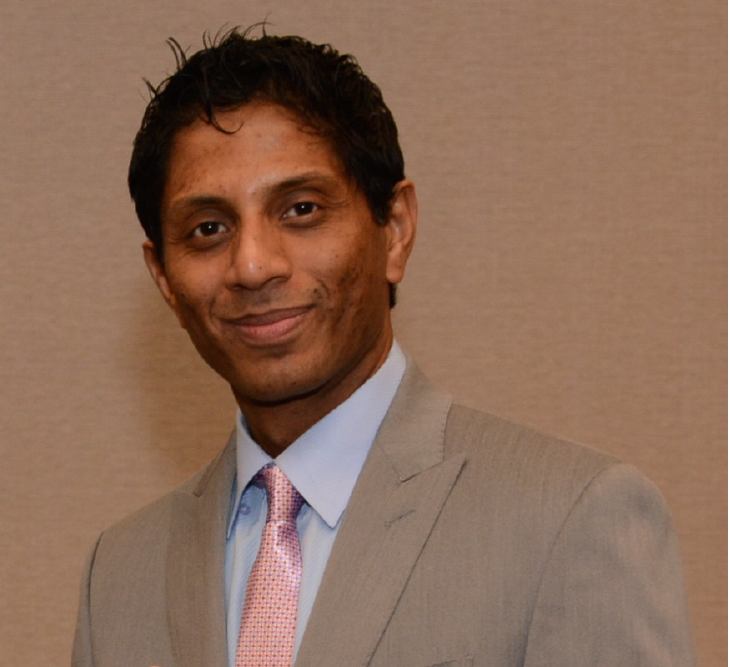}}]{Shrikanth (Shri) Narayanan} (StM'88-M'95-SM'02-F'09) is the Niki \& C. L. Max Nikias Chair in Engineering at the University of Southern California (USC), and holds appointments as Professor of Electrical and Computer Engineering, Computer Science, Linguistics, Psychology, Neuroscience, Otolaryngology and Pediatrics, Research Director of the Information Science Institute, and director of the Ming Hsieh Institute. Prior to USC he was with AT\&T Bell Labs and AT\&T Research from 1995-2000. At USC, he directs the Signal Analysis and Interpretation Laboratory (SAIL). His research focuses on human-centered signal and information processing and systems modeling with an interdisciplinary emphasis on speech, audio, language, multimodal and biosignal processing and machine intelligence, and their applications with direct societal relevance. [http://sail.usc.edu]

Prof. Narayanan is a Fellow of the National Academy of Inventors, the Acoustical Society of America, IEEE, the International Speech Communication Association (ISCA), the Association for Psychological Science, the American Institute for Medical and Biological Engineering (AIMBE), and the American Association for the Advancement of Science (AAAS) and a member of Tau Beta Pi, Phi Kappa Phi, and Eta Kappa Nu. He is VP-Education for IEEE Signal Processing Society, an Editor for the Computer Speech and Language Journal and an Associate Editor for the APSIPA Transactions on Signal and Information Processing. He was also previously Editor in Chief for IEEE Journal of Selected Topics in Signal Processing and an Associate Editor of the IEEE Transactions on Speech and Audio Processing (2000–2004), IEEE Signal Processing Magazine (2005–2008), IEEE Transactions on Multimedia (2008-2011), IEEE Transactions on Signal and Information Processing over Networks (2014-2015), IEEE Transactions on Affective Computing (2010-2016), and the Journal of the Acoustical Society of America (2009-2017). He is a recipient of several honors including the 2015 Engineers Council’s Distinguished Educator Award, a Mellon award for mentoring excellence, the 2005 and 2009 Best Journal Paper awards from the IEEE Signal Processing Society and serving as its Distinguished Lecturer for 2010-11, a 2018 ISCA Best Journal Paper award, and serving as an ISCA Distinguished Lecturer for 2015-16 and the Willard R. Zemlin Memorial Lecturer for ASHA in 2017. Papers co-authored with his students have won awards including the 2014 Ten-year Technical Impact Award from ACM ICMI and at several conferences. He has published over 800 papers and has been granted seventeen U.S. patents.
\end{IEEEbiography}

\clearpage
\begin{figure*}
        \centering
        \begin{subfigure}{\textwidth}
        \centering
        \begin{subfigure}{0.2\textwidth}
            \centering
            \includegraphics[scale=0.4]{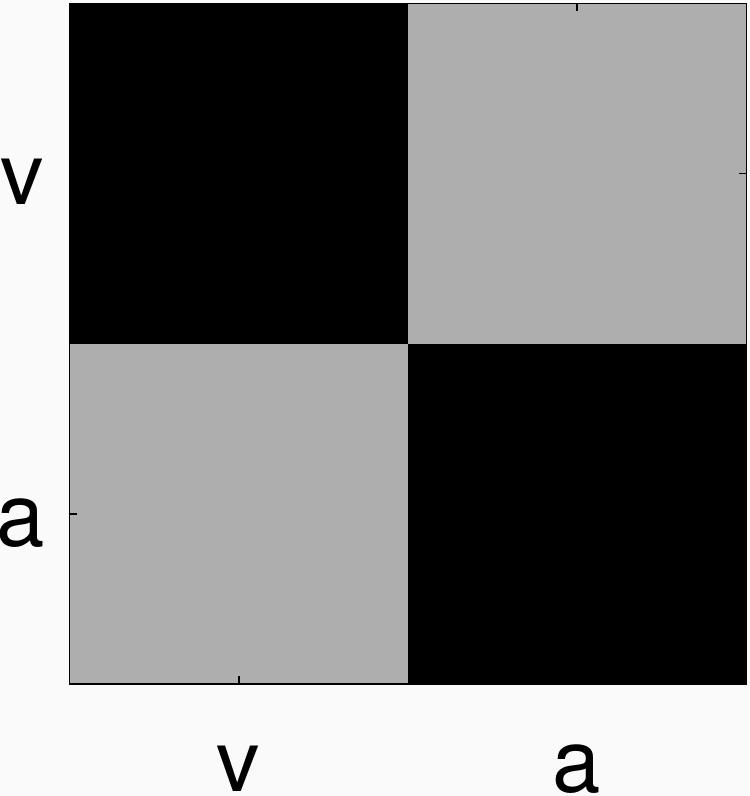}
        \end{subfigure}
        \begin{subfigure}{0.2\textwidth}
            \centering
            \includegraphics[scale=0.4]{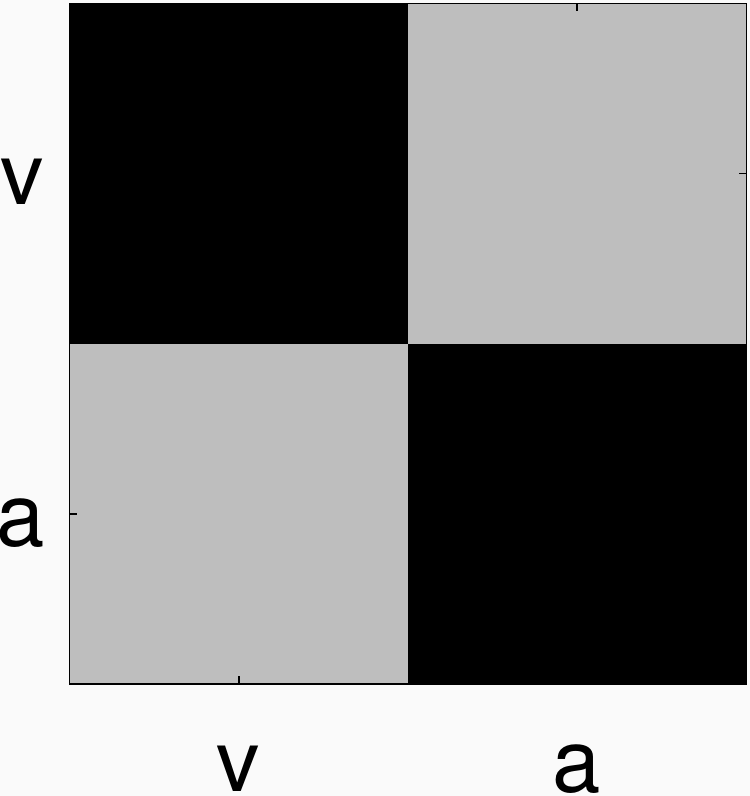}
        \end{subfigure}
        \begin{subfigure}{0.2\textwidth}
            \centering
            \includegraphics[scale=0.4]{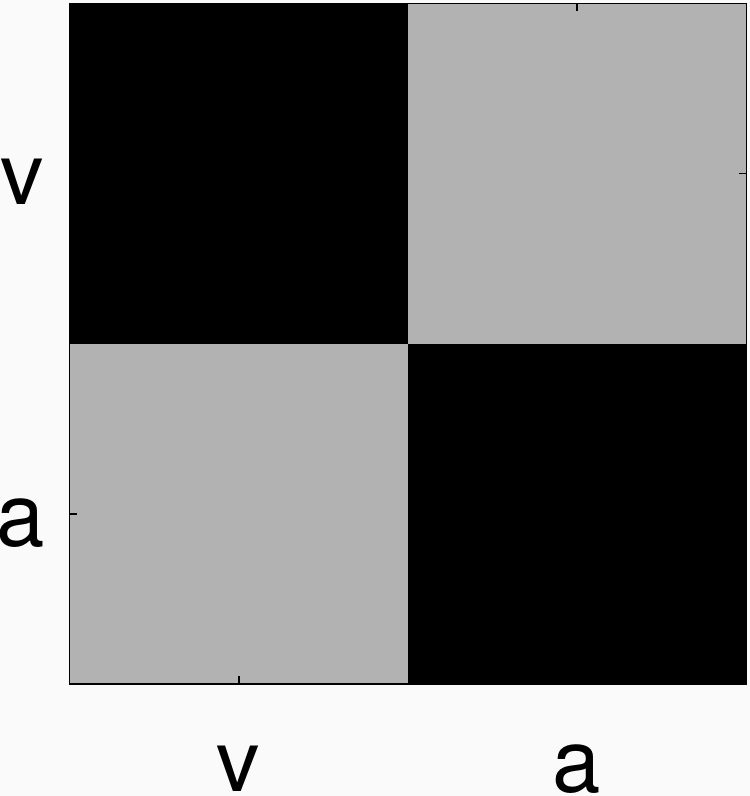}
        \end{subfigure}
        \begin{subfigure}{0.2\textwidth}
            \centering
            \includegraphics[scale=0.4]{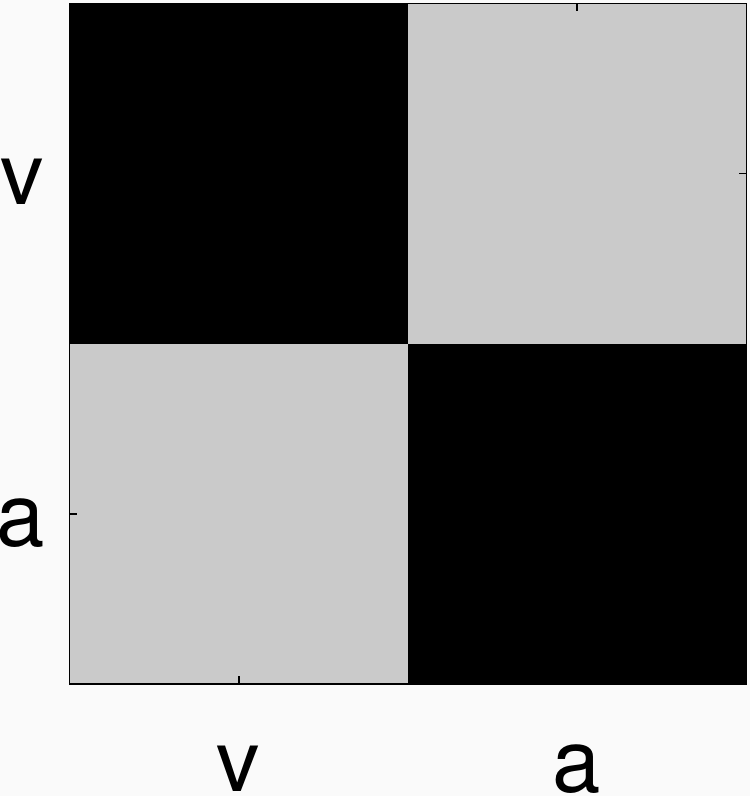}
        \end{subfigure}
        \caption{Movie emotions corpus}
        \end{subfigure}
        \par\medskip
        \begin{subfigure}{\textwidth}
        \centering
        \begin{subfigure}{0.2\textwidth}
            \centering
            \includegraphics[scale=0.4]{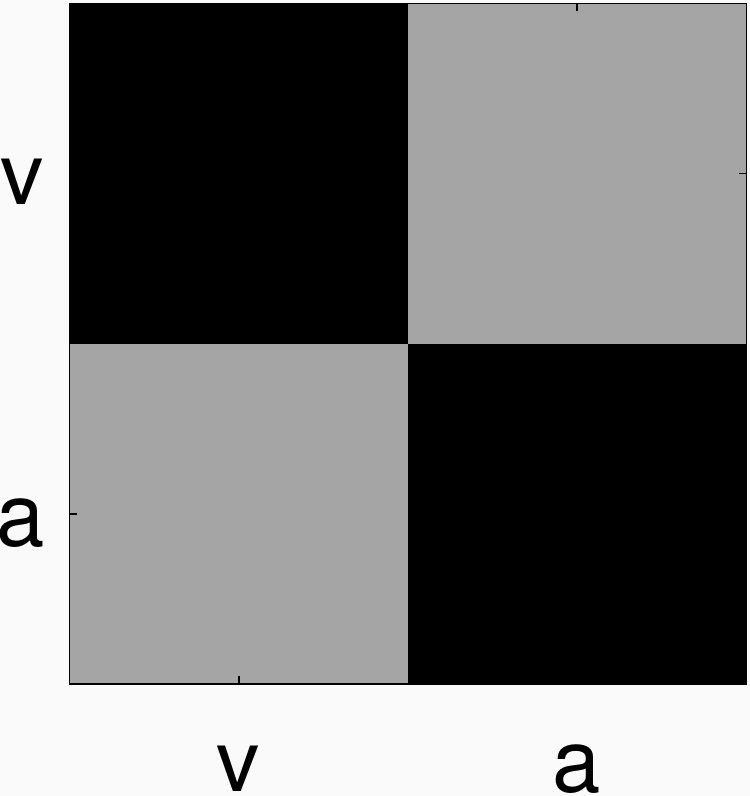}
        \end{subfigure}
        \begin{subfigure}{0.2\textwidth}
            \centering
            \includegraphics[scale=0.4]{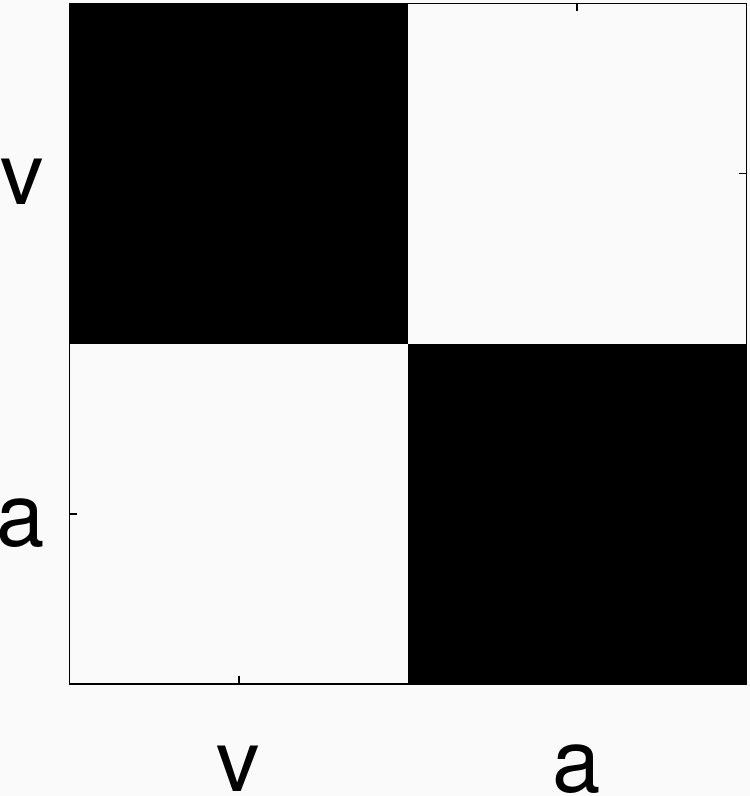}
        \end{subfigure}
        \begin{subfigure}{0.2\textwidth}
            \centering
            \includegraphics[scale=0.4]{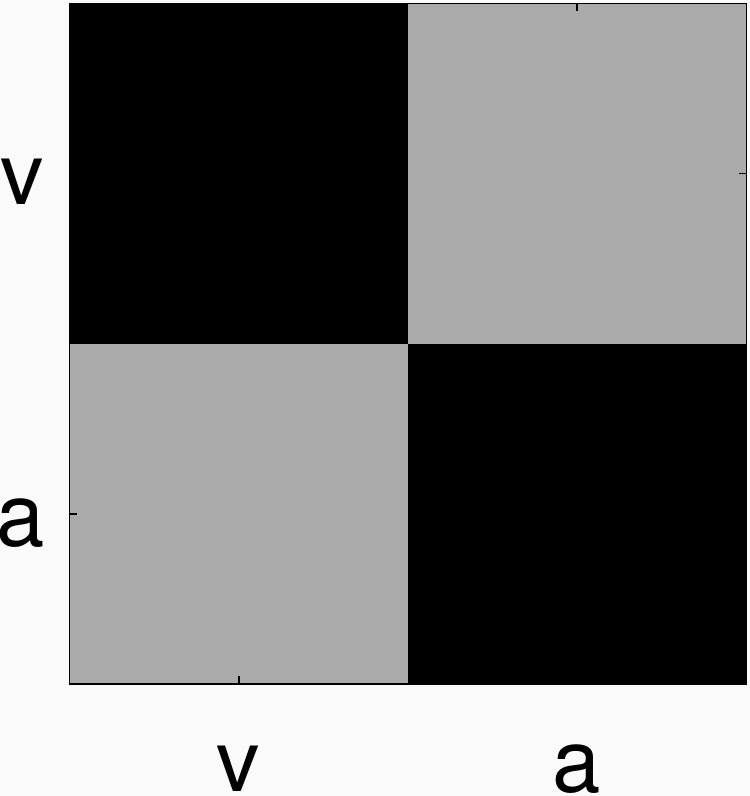}
        \end{subfigure}
        \begin{subfigure}{0.2\textwidth}
            \centering
            \includegraphics[scale=0.4]{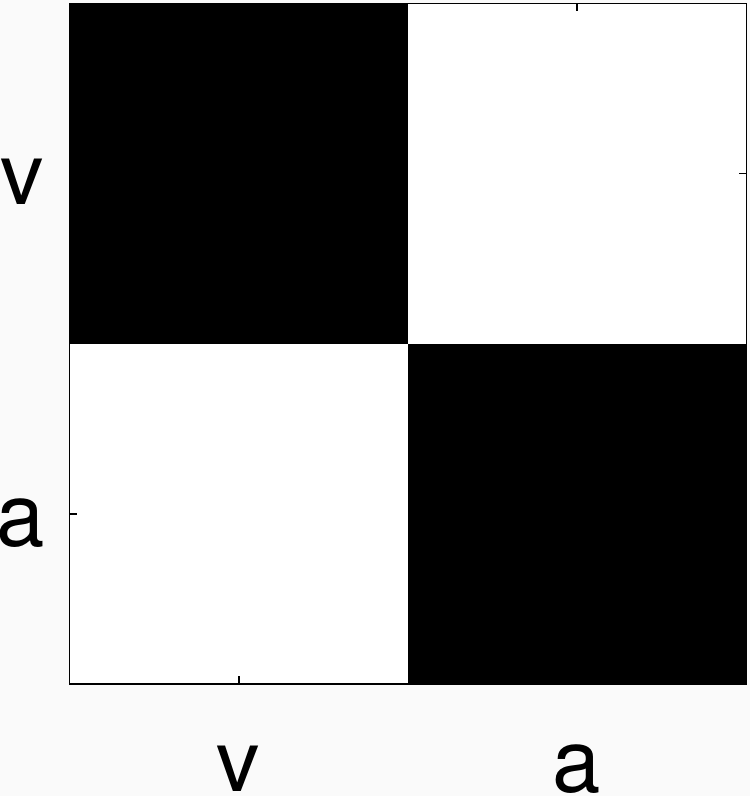}
        \end{subfigure}
        \subcaption{RECOLA}
        \par\medskip
        \end{subfigure}
        \begin{subfigure}{\textwidth}
           \centering 
           \includegraphics[scale=0.7]{colorbar}
        \end{subfigure}
        \caption{Annotator specific correlations between annotation dimensions for the Movie emotions and RECOLA corpora; v-valence, a-arousal}
        \label{fig:annot_corrs}
\end{figure*}

\begin{appendices}
\section{Supplementary Analyses}
\label{appendix_analysis}

\subsection{Annotator specific correlations}
\label{appendix_analysis_corrs}

Figure \ref{fig:annot_corrs} highlights the correlations between annotation dimensions for 4 annotators from the movie emotions \cite{malandrakis2011supervised} and RECOLA \cite{ringeval2013introducing} corpora. As noted earlier, different annotators may exhibit different degree of associations between the annotation dimensions, leading to the observed differences in correlations both within and between the two corpora. This difference in annotator behavior also leads to the different inter-dimension correlations observed among the corpora in Figure \ref{fig:corrs}.

\subsection{Effect of dependency among dimensions}
\label{appendix_analysis_fk}
\begin{figure*}
        \centering
        \begin{subfigure}{0.3\textwidth}
            \centering
            \includegraphics[scale=0.5]{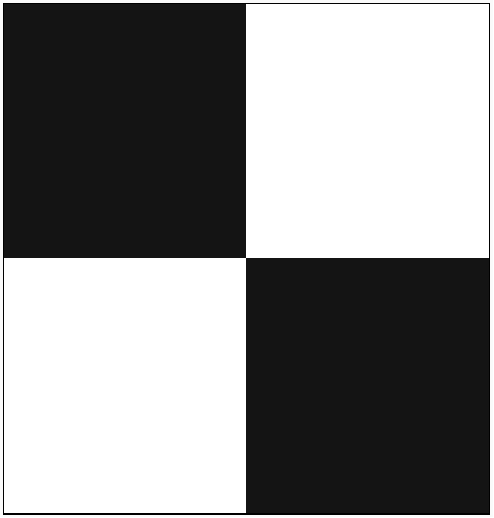}
            \caption{0}
        \end{subfigure}
        \begin{subfigure}{0.3\textwidth}
            \centering
            \includegraphics[scale=0.5]{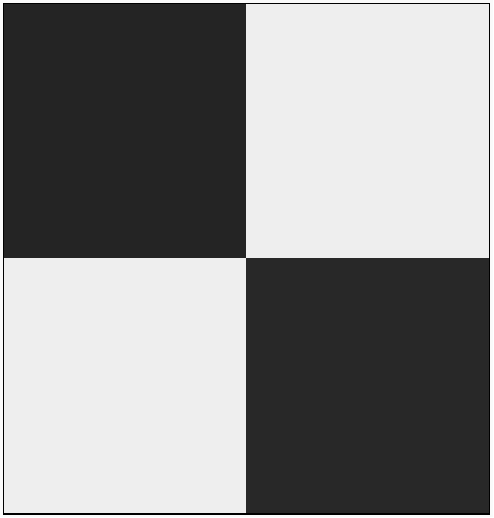}
            \caption{0.2}
        \end{subfigure}
        \begin{subfigure}{0.3\textwidth}
            \centering
            \includegraphics[scale=0.5]{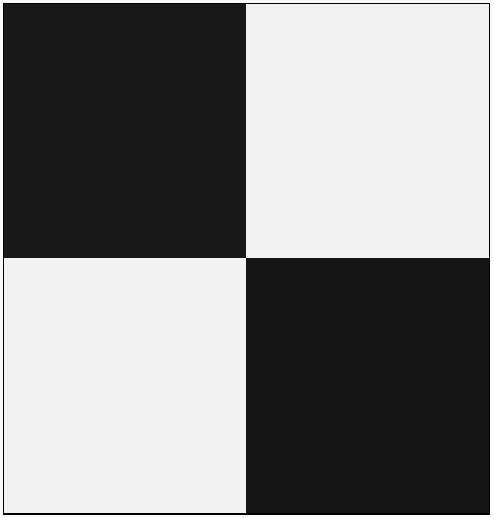}
            \caption{0.4}
        \end{subfigure}
        \par\medskip
        \begin{subfigure}{0.3\textwidth}
            \centering
            \includegraphics[scale=0.5]{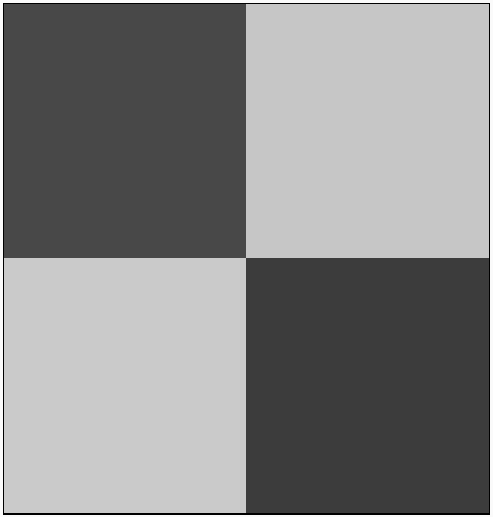}
            \caption{0.6}
        \end{subfigure}
        \begin{subfigure}{0.3\textwidth}
            \centering
            \includegraphics[scale=0.5]{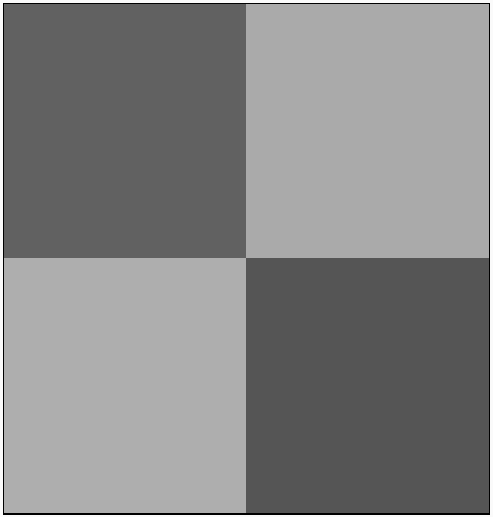}
            \caption{0.8}
        \end{subfigure}
        \begin{subfigure}{0.3\textwidth}
            \centering
            \includegraphics[scale=0.5]{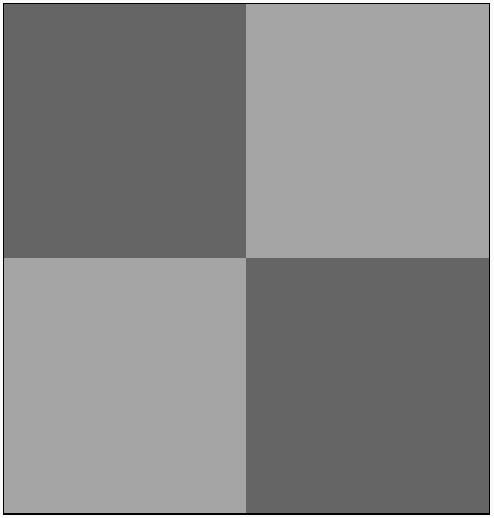}
            \caption{1}
            \label{fig:disc_dependency_Fk_f}
        \end{subfigure}
        \par\medskip
        \begin{subfigure}{\textwidth}
           \centering 
           \includegraphics[scale=0.7]{colorbar}
        \end{subfigure}
        \caption{Average $F_k$ plots estimated from the joint model at different step sizes for off diagonal elements of the annotator's $F_k$ matrices}
        \label{fig:disc_dependency_Fk}
\end{figure*}

The model we present includes the annotator specific parameter $F_k$ which measures the relationships between the annotation dimensions. To highlight the ability of the model to recover this parameter, in Figure \ref{fig:disc_dependency_Fk}, we show a plot of averages of all predicted $F_k$ matrices for different step sizes from the synthetic experiment described in Section \ref{subsec:varying_fk}. 
In each case, the predicted $F_k$ matrices closely resemble the actual matrices for the annotators highlighting the accuracy of the joint model. However, as we get closer to step size $1$, the estimated $F_k$ matrices appear to be washed out (despite being accurate to a scaling term), with all terms of the estimated $F_k$ close to $0.5$ instead of $1$ (Figure \ref{fig:disc_dependency_Fk_f}), due to model unidentifiability. 

\section{EM derivation for global annotation model}
\label{appendix_global}
\subsubsection{Deriving $p(\textbf{a}^m_1\dots\textbf{a}^m_K, \textbf{a}^m_*)$}

To help with the model formulation, we first derive parameters of the joint distribution $p(\textbf{a}^m_1\dots\textbf{a}^m_K, \textbf{a}^m_*)$. Since the product of two normal distributions is also normal \cite{bishop2006pattern}, this joint distribution is also normal and is given by, 

\begin{align}
\begin{bmatrix}\textbf{a}^m_*\\
\textbf{a}^m_1\\
\vdots\\
\textbf{a}^m_K
\end{bmatrix} &\sim  N
\begin{pmatrix}
\begin{bmatrix}
\Theta^T \textbf{x}_m\\
F_1\Theta^T \textbf{x}_m\\
\vdots\\
F_K\Theta^T \textbf{x}_m
\end{bmatrix}\!\!,&
\begin{bmatrix}
\Sigma_{**} & \Sigma_{*1} & \dots & \Sigma_{*K}\\
\Sigma_{1*} & \Sigma_{11} & \dots & \Sigma_{1K}\\
\vdots & \vdots & \ddots & \vdots\\
\Sigma_{K*} & \Sigma_{K1} & \dots & \Sigma_{KK} 
\end{bmatrix} 
\end{pmatrix}
\label{label:joint_dist} 
\end{align}

The different components of the covariance matrix from Equation \ref{label:joint_dist} are derived below.
\begin{align}
\Sigma_{**} &= Cov(\textbf{a}^m_*) \nonumber \\
&= \sigma_*^2I \nonumber \\
\Sigma_{k*} &= \E [\textbf{a}^m_k (\textbf{a}^m_*)^T] - \E[\textbf{a}^m_k ]\E[(\textbf{a}^m_*)^T] \nonumber \\
&= \E[(F_k \textbf{a}^m_* + \bm{\eta}_k) (\textbf{a}^m_*)^T] - \E[F_k \textbf{a}^m_* + \bm{\eta}_k]\E[(\textbf{a}^m_*)^T] \nonumber \\
&= F_k(\sigma_*^2I) \nonumber \\ 
\Sigma_{kk} &= Cov(F_k \textbf{a}^m_* +  \bm{\eta}_k) \nonumber \\
&= Cov(F_k \textbf{a}^m_*) + \tau_k^2 I \nonumber \\
&= F_k\Sigma_{**}F_k^T + \tau_k^2 I  \nonumber \\
&= \sigma_*^2F_kF_k^T + \tau_k^2 I  \nonumber \\
\Sigma_{k_ik_j} &= \E_{\textbf{a}^m_*}[Cov(\textbf{a}^m_{k_1},\textbf{a}^m_{k_2}|\textbf{a}^m_*)] + Cov(\E[\textbf{a}^m_{k_1}|\textbf{a}^m_*], \E[\textbf{a}^m_{k_2}|\textbf{a}^m_*]) \nonumber \\
&= Cov(\E[\textbf{a}^m_{k_1}|\textbf{a}^m_*], \E[\textbf{a}^m_{k_2}|\textbf{a}^m_*]) \nonumber \\
&= Cov(F_{k_1}\textbf{a}^m_*, F_{k_2}\textbf{a}^m_*) \nonumber \\
&= F_{k_1}\Sigma_{**}(F_{k_2})^T \nonumber \\
&= \sigma^2_*F_{k_1}F_{k_2}^T \nonumber
\end{align}

In the derivation of $\Sigma_{k_ik_j}$, the first equation is a direct application of the law of total covariance and the second equation is because of the conditional independence assumption of annotation values $\textbf{a}^m_{k_i}$ given the ground truth $\textbf{a}^m_{*}$

Finally, owing to the jointly normal distributions, $p(\textbf{a}^m_{*} | \textbf{a}^m_1 \dots \textbf{a}^m_K)$ is also normal:
\begin{align}
p(\textbf{a}^m_{*} | \textbf{a}^m_1 \dots \textbf{a}^m_K) \sim N(\mu_{\textbf{a}^m_{*} | \textbf{a}^m_1 \dots \textbf{a}^m_K} | \Sigma_{\textbf{a}^m_{*} | \textbf{a}^m_1 \dots \textbf{a}^m_K} ) \nonumber
\end{align}

Also, by definitions of conditional normal distributions, given a normal vector of the form 
\begin{align}
\begin{bmatrix}
x_1\\
x_2
\end{bmatrix}
\sim N
\begin{pmatrix}
\begin{bmatrix}
\mu_1\\
\mu_2
\end{bmatrix}\!\!,&
\begin{bmatrix}
\Sigma_{11} & \Sigma_{12} \\
\Sigma_{21} & \Sigma_{22} \\
\end{bmatrix}
\end{pmatrix} \nonumber
\end{align}

the conditional distribution $p(x_1|x_2) \sim N(\mu_{x_1|x_2}, \Sigma_{x_1|x_2})$ has the following form. 
\begin{align}
\label{condnormala}
\mu_{x_1|x_2} &= \mu_1 + \Sigma_{12}\Sigma_{22}^{-1}(x_2-\mu_2) \\
\label{condnormalb}
\Sigma_{x_1|x_2} &= \Sigma_{11} - \Sigma_{12}\Sigma_{22}^{-1}\Sigma_{21} 
\end{align}
\subsection{EM Formulation}
\label{discreteem}
We begin by introducing a new distribution $q(\textbf{a}^m_*)$ in Equation \ref{eqn:model_likelihood}. We drop the parameters $\Phi$ from the likelihood function expansion for convenience.  

\begin{equation}
\log \mathcal{L} = \sum_{m=1}^M \log \int_{\textbf{a}^m_*} q(\textbf{a}^m_*) \frac{p(\textbf{a}^m_1\dots\textbf{a}^m_K | \textbf{a}^m_*) p(\textbf{a}^m_*)}{q(\textbf{a}^m_*)}  \, d\textbf{a}^m_* 
\end{equation}

Using Jensen's inequality over log of expectation, we can write the above as follows,

\begin{equation}
\log \mathcal{L} \geq \sum_{m=1}^M \int_{\textbf{a}^m_*} q(\textbf{a}^m_*) \log \frac{p(\textbf{a}^m_1\dots\textbf{a}^m_K | \textbf{a}^m_*) p(\textbf{a}^m_*)}{q(\textbf{a}^m_*)}  \, d\textbf{a}^m_* 
\label{lowerboundl}
\end{equation}

The bound above becomes tight when the expectation is taken over a constant value, i.e.

\begin{gather}
\frac{p(\textbf{a}^m_1\dots\textbf{a}^m_K | \textbf{a}^m_*) p(\textbf{a}^m_*)}{q(\textbf{a}^m_*)} = c \nonumber
\end{gather}

Solving for the constant c, we have
\begin{align}
q(\textbf{a}^m_*) = \frac{p(\textbf{a}^m_1\dots\textbf{a}^m_K, \textbf{a}^m_*)}{p(\textbf{a}^m_1\dots\textbf{a}^m_K)} = p(\textbf{a}^m_* | \textbf{a}^m_1\dots\textbf{a}^m_K) \nonumber
\end{align} 

\subsubsection{E-Step}

The E-step involves simply assuming $q(\textbf{a}^m_*)$ to follow the conditional distribution $p(\textbf{a}^m_* | \textbf{a}^m_1\dots\textbf{a}^m_K)$. 

To help with future computations, we also compute the following expectations, where the first two are a result of equations \ref{condnormala} and \ref{condnormalb}; third equation is by definition of covariance and the last one is a standard result (see  the matrix cookbook eq. 327). 
\begin{align}
\E_{\textbf{a}^m_* | \textbf{a}^m_1\dots\textbf{a}^m_K}[\textbf{a}^m_*] &= \Theta^T \textbf{x}_m + \Sigma_{\textbf{a}^m_*,\textbf{a}^m_1\dots\textbf{a}^m_K}\nonumber\\&(\Sigma_{\textbf{a}^m_1\dots\textbf{a}^m_K,\textbf{a}^m_1\dots\textbf{a}^m_K})^{-1}(\textbf{a}^m - \boldsymbol{\mu}^m) \nonumber \\
\Sigma_{\textbf{a}^m_* | \textbf{a}^m_1\dots\textbf{a}^m_K} [\textbf{a}^m_*] &= \Sigma_{\textbf{a}^m_*,\textbf{a}^m_*} - \Sigma_{\textbf{a}^m_*,\textbf{a}^m_1\dots\textbf{a}^m_K}\nonumber\\&(\Sigma_{\textbf{a}^m_1\dots\textbf{a}^m_K,\textbf{a}^m_1\dots\textbf{a}^m_K})^{-1}\Sigma_{\textbf{a}^m_1\dots\textbf{a}^m_K,\textbf{a}^m_*} \nonumber \\
\E_{\textbf{a}^m_* | \textbf{a}^m_1\dots\textbf{a}^m_K}  [\textbf{a}^m_* (\textbf{a}^m_* )^T] &= \Sigma_{\textbf{a}^m_* | \textbf{a}^m_1\dots\textbf{a}^m_K} [\textbf{a}^m_*] + \nonumber \\ & \E_{\textbf{a}^m_* | \textbf{a}^m_1\dots\textbf{a}^m_K}[\textbf{a}^m_*] \E_{\textbf{a}^m_* | \textbf{a}^m_1\dots\textbf{a}^m_K}[(\textbf{a}^m_*)^T] \nonumber \\
\E_{\textbf{a}^m_* | \textbf{a}^m_1\dots\textbf{a}^m_K}  [(\textbf{a}^m_*)^T\textbf{a}^m_*  ] &= trace(\Sigma_{\textbf{a}^m_* | \textbf{a}^m_1\dots\textbf{a}^m_K} [\textbf{a}^m_*]) + \nonumber \\ &\E_{\textbf{a}^m_* | \textbf{a}^m_1\dots\textbf{a}^m_K}[(\textbf{a}^m_*)^T] \E_{\textbf{a}^m_* | \textbf{a}^m_1\dots\textbf{a}^m_K}[\textbf{a}^m_*] \nonumber
\end{align}

$\textbf{a}^m$ and $\boldsymbol{\mu}^m$ are $DK$ dimensional vectors obtained by concatenating the $K$ annotation vectors $\textbf{a}^m_1,\dots\textbf{a}^m_K$ and their corresponding expected values $F_1\Theta^T \textbf{x}_m\dots F_K\Theta^T \textbf{x}_m$.

\subsubsection{M-step}
In the M-step, we find the parameters of the model by maximizing Equation \ref{lowerboundl}. We first write this equation as an expectation and an equality. The expectation below is with respect to $q(\textbf{a}^m_*) = p(\textbf{a}^m_* | \textbf{a}^m_1\dots\textbf{a}^m_K)$; we drop the subscript for ease of exposition
\begin{align}
\log \mathcal{L} &= \sum_{m=1}^M \E_{\textbf{a}^m_* | \textbf{a}^m_1\dots\textbf{a}^m_K} \big[ \log \frac{p(\textbf{a}^m_1\dots\textbf{a}^m_K | \textbf{a}^m_*) p(\textbf{a}^m_*)}{q(\textbf{a}^m_*)}  \big] \nonumber  \\
\log \mathcal{L} &= \sum_{m=1}^M \E \log p(\textbf{a}^m_1\dots\textbf{a}^m_K | \textbf{a}^m_*) + \E \log p(\textbf{a}^m_*) + \mathcal{H} \nonumber \\
\label{eqn:objective}
\log \mathcal{L} &= \sum_{m=1}^M \big( \sum_{k=1}^K \E \log p(\textbf{a}^m_k | \textbf{a}^m_*) + \E \log p(\textbf{a}^m_*) + \mathcal{H} \big)
\end{align}

where $p(\textbf{a}^m_*)$ and $p(\textbf{a}^m_k | \textbf{a}^m_*)$ are given by equations \ref{eqn:a_star_discrete} and \ref{eqn:ann_discrete} respectively. The last equation above uses that fact that we assume independence among annotators given the ground truth. Also expectation commutes with the linear sum over the $K$ terms.

Here, $\mathcal{H}$ is the entropy of $p(\textbf{a}^m_* | \textbf{a}^m_1\dots\textbf{a}^m_K)$. We maximize Equation \ref{eqn:objective} with respect to each of the parameters to obtain the M-step updates.

\textbf{Estimating $\bm{F}_k$}
Differentiating Equation (\ref{eqn:objective}) with respect to $F_k$ and equating the derivative to 0
\begin{gather*}
\Delta_{F_k} Q = 0 \nonumber \\
\Delta_{F_k} \sum_{m=1}^{M_k} \E [(\textbf{a}^m_k - F_k\textbf{a}^m_*)^T(\tau_k^2I)^{-1}(\textbf{a}^m_k - F_k\textbf{a}^m_*)] = 0 \nonumber \\
\Delta_{F_k} \frac{1}{\tau_k^2}\sum_{m=1}^{M_k} \E [(\textbf{a}^m_k - F_k\textbf{a}^m_*)^T(\textbf{a}^m_k - F_k\textbf{a}^m_*)] = 0 \nonumber \\
\sum_{m=1}^{M_k} - 2\textbf{a}^m_k\E[(\textbf{a}^m_*)^T] + 2F_k\E[\textbf{a}^m_*(\textbf{a}^m_*)^T] = 0 \nonumber \\
\therefore F_k = \bigg( \sum_{m=1}^{M_k} \textbf{a}^m_k \E[(\textbf{a}^m_*)^T] \bigg) \bigg( \sum_{m=1}^{M_k}  \E[\textbf{a}^m_*(\textbf{a}^m_*)^T]\bigg)^{-1}
\end{gather*}

where, $M_k$ is the number of points annotated by user $k$.

We used the following facts in the above derivation: $trace(x) = x$ for scalar x; $trace(AB) = trace(BA)$; $\Delta_A trace(A^Tx) = x$ and $\Delta_A trace(A^TAB) = AB+AB^T$ for matrix $A$. We also make use of the fact that expectation and trace of a matrix are commutative since trace is a linear sum. 

\textbf{Estimating $\bm{\Theta}$} Similarly, to find $\Theta$, we differentiate Equation (\ref{eqn:objective}) with respect to $\Theta$ and equate it to 0.
\begin{gather*}
\Delta_\Theta Q = 0 \nonumber \\
\Delta_\Theta \sum_{m=1}^M \E [(\textbf{a}^m_* - \Theta^T \textbf{x}_m)^T(\sigma^2I)^{-1}(\textbf{a}^m_* - \Theta^T \textbf{x}_m)] = 0 \nonumber \\
\Delta_{\Theta} \frac{1}{\sigma^2} \sum_{m=1}^M \E [(\textbf{a}^m_* - \Theta^T \textbf{x}_m)^T(\textbf{a}^m_* - \Theta^T \textbf{x}_m)] = 0 \nonumber \\
\sum_{m=1}^M -2 \textbf{x}_m\E[(\textbf{a}^m_*)^T] + 2  \textbf{x}_m\textbf{x}_m^T\Theta = 0 \nonumber \\
\Theta = \bigg( \sum_{m=1}^M \textbf{x}_m \textbf{x}^T_m \bigg)^{-1} \bigg(\sum_{m=1}^M \textbf{x}_m \E[(\textbf{a}^m_*)^T]\bigg) \nonumber \\
\therefore \Theta = (\textrm{X}^T\textrm{X})^{-1}(\textrm{X}^T\E[\textbf{a}^m_*])
\end{gather*}
which looks like the familiar normal equation except we use the expected value of $\textbf{a}_*$. Here, $\textrm{X}$ is the matrix of features of the $M$ data instances; it includes individual feature vectors $x_m$ in its rows.

\textbf{Estimating $\bm{\sigma}$}
Differentiating Equation (\ref{eqn:objective}) with respect to $\sigma$ and equating to 0, we have 

\begin{gather*}
\Delta_{\sigma} Q = 0 \nonumber \\
\Delta_{\sigma} \sum_{m=1}^{M}\bigg(-D\log \sigma - \frac{1}{2\sigma^2} \big(\E[(\textbf{a}^m_*)^T\textbf{a}^m_*]-   2tr(\Theta^T\textbf{x}_m\E[(\textbf{a}^m_*)^T]) \\ + tr(\textbf{x}^T_m\Theta\Theta^T\textbf{x}_m) \big) \bigg) = 0 \nonumber \\ 
\sum_{m=1}^{M} - \frac{D}{\sigma} + \frac{1}{\sigma^3} \bigg( \E[(\textbf{a}^m_*)^T\textbf{a}^m_*] -   2tr(\Theta^T\textbf{x}_m\E[(\textbf{a}^m_*)^T]) + \\ tr(\textbf{x}^T_m\Theta\Theta^T\textbf{x}_m) \bigg)  = 0 \nonumber \\
\frac{MD}{\sigma} = \frac{1}{\sigma^3} \sum_{m=1}^{M}\bigg( \E[(\textbf{a}^m_*)^T\textbf{a}^m_*] -   2tr\big(\Theta^T\textbf{x}_m\E[(\textbf{a}^m_*)^T]\big) + \\ tr(\textbf{x}^T_m\Theta\Theta^T\textbf{x}_m) \bigg)  \nonumber \\
\therefore \sigma^2 = \frac{1}{MD} \sum_{m=1}^{M} \bigg( \E[(\textbf{a}^m_*)^T\textbf{a}^m_*] -   2tr\big(\Theta^T\textbf{x}_m\E[(\textbf{a}^m_*)^T]\big) + \\ tr(\textbf{x}^T_m\Theta\Theta^T\textbf{x}_m) \bigg)\nonumber
\end{gather*}

\textbf{Estimating $\bm{\tau_k}$}
Differentiating Equation (\ref{eqn:objective}) with respect to $\tau_k$ and equating to 0, we have 

\begin{gather*}
\Delta_{\tau_k} Q = 0 \nonumber \\
\Delta_{\tau_k} \sum_{m=1}^{M_k} \bigg( -D \log \tau_k - \frac{1}{2\tau_k^2} \big( (\textbf{a}^m_k)^T\textbf{a}^m_k - 2tr(F_k^T \textbf{a}^m_k\E[(\textbf{a}^m_*)^T]) + \\ tr(F_k^TF_k\E[\textbf{a}^m_*(\textbf{a}^m_*)^T]) \big) \bigg) = 0 \nonumber \\
\sum_{m=1}^{M_k} \bigg( - \frac{D}{\tau_k} + \frac{1}{\tau_k^3} \big( (\textbf{a}^m_k)^T\textbf{a}^m_k - 2tr(F_k^T \textbf{a}^m_k\E[(\textbf{a}^m_*)^T]) + \\ tr(F_k^TF_k\E[\textbf{a}^m_*(\textbf{a}^m_*)^T])  \big) \bigg) = 0 \nonumber \\
\therefore \tau_k^2 = \frac{1}{DM_k} \sum_{m=1}^{M_k} \bigg( (\textbf{a}^m_k)^T\textbf{a}^m_k - 2tr(F_k^T \textbf{a}^m_k\E[(\textbf{a}^m_*)^T]) + \\ tr(F_k^TF_k\E[\textbf{a}^m_*(\textbf{a}^m_*)^T])  \bigg)
\end{gather*}

\section{EM derivation for 
time series annotation model}
\label{appendix_timeseries}
\subsection{EM Formulation}
Similar to the process described in Appendix \ref{appendix_global}, the log likelihood function for the time series model is shown below (similar to Equation \ref{lowerboundl}).

\begin{align}
\log \mathcal{L} \geq \sum_{m=1}^M \int_{\textbf{a}^m_*} q(\textbf{a}^m_*) \log \frac{p(\textbf{a}^m_1\dots\textbf{a}^m_K | \textbf{a}^m_*) p(\textbf{a}^m_*)}{q(\textbf{a}^m_*)}  \, d\textbf{a}^m_* 
\label{continuousll}
\end{align}

The bound becomes tight when $q(\textbf{a}^m_*) = p(\textbf{a}^m_* | \textbf{a}^m_1\dots\textbf{a}^m_K)$.

\subsubsection{E-step}

Computing the expectation function over the entire distribution of $q(\textbf{a}^m_*)$ is computationally expensive since $\textbf{a}^m_*$ is a matrix. To avoid this, we instead use \textit{Hard-EM} in which we assume a dirac-delta distribution for $\textbf{a}^m_*$ which is centered at the mode of $q(\textbf{a}^m_*)$. This is a common practice in latent models and is the approach followed by \cite{Gupta2015} in estimating the annotator filter parameters. We assign this value to $\textbf{a}^m_*$ in the E-step:
\begin{align}
\textbf{a}^m_* &= \argmax_{\textbf{a}^m_*} q(\textbf{a}^m_*) \nonumber \\
 &= \argmax_{\textbf{a}^m_*} p(\textbf{a}^m_* |\textbf{a}^m_1,\dots\textbf{a}^m_K) \nonumber \\
 &= \argmax_{\textbf{a}^m_*} \frac{p(\textbf{a}^m_*, \textbf{a}^m_1,\dots\textbf{a}^m_K)}{p(\textbf{a}^m_1,\dots\textbf{a}^m_K)} \nonumber \\
 &= \argmax_{\textbf{a}^m_*} p(\textbf{a}^m_1,\dots\textbf{a}^m_K | \textbf{a}^m_*)p(\textbf{a}^m_*) \nonumber \\
 &= \argmax_{\textbf{a}^m_*} \log p(\textbf{a}^m_1,\dots\textbf{a}^m_K| \textbf{a}^m_*)p(\textbf{a}^m_*) \nonumber \\
\textbf{a}^m_* &= \argmax_{\textbf{a}^m_*} \big( \log p(\textbf{a}^m_1,\dots\textbf{a}^m_K | \textbf{a}^m_*) + \log p(\textbf{a}^m_*| \textbf{x}_m) \big) \nonumber 
\end{align}

Since we assume that each annotator is independent of the others given the ground truth, we have

\begin{align}
\textbf{a}^m_* &= \argmax_{\textbf{a}^m_*} \log \prod_k p(\textbf{a}^m_k | \textbf{a}^m_*) + \log p(\textbf{a}^m_*) \nonumber \\
\textbf{a}^m_* &= \argmax_{\textbf{a}^m_*} \sum_k \log p(\textbf{a}^m_k | \textbf{a}^m_*) + \log p(\textbf{a}^m_*) \nonumber 
\end{align}

Further, since each annotation dimension $\textbf{a}^{m,d}_k$ is assumed to independent given $\textbf{a}^m_*$, we have

\begin{align}
\textbf{a}^m_* &= \argmax_{\textbf{a}^m_*} \sum_k \sum_d \log p(\textbf{a}^{m,d}_k | \textbf{a}^m_*) + \log p(\textbf{a}^m_*) \nonumber
\end{align}

Finally, since both $\textbf{a}^{m,d}_k$ and $\textbf{a}^m_*$ are defined using IID Gaussian noise, the above maximization problem is equivalent to the following minimization. 

\begin{gather}
\textbf{a}^m_* = \argmin_{\textbf{a}^m_*} \sum_k \sum_d ||\textbf{a}^{m,d}_k - F^d_k \text{vec}(\textbf{a}^m_*)||^2_2 + \nonumber \\ ||\text{vec}(\textbf{a}^m_*) - \text{vec}(\textrm{X}_m\Theta)||^2_2 \nonumber
\end{gather}

For convenience, we reshape $\textbf{a}^m_*$ into a vector and optimize with respect to the flattened vector. If we choose \text{vec}($\textbf{a}^m_*) = v$ and  $\text{vec}(\textrm{X}_m\Theta) = y$, the objective becomes,

\begin{align}
Q(v) = \sum_k \sum_d ||\textbf{a}^{m,d}_k - F^d_k v||^2_2 + ||v - y||^2_2 \nonumber
\end{align}

Differentiating $Q$ with respect to $v$ and equating the gradient to 0, we get

\begin{gather}
\Delta_v Q = 0 \nonumber \\
\Delta_v \sum_k \sum_d (\textbf{a}^{m,d}_k - F^d_k v)^T (\textbf{a}^{m,d}_k - F^d_k v) + \nonumber \\ (v-y)^T(v-y) = 0 \nonumber \\
\Delta_v \sum_k \sum_d (\textbf{a}^{m,d}_k)^T\textbf{a}^{m,d}_k + v^T(F^d_k)^T F^d_k v - \nonumber \\ 2(\textbf{a}^{m,d}_k)^T F^d_kv  + (v^Tv-2y^Tv+y^Ty) = 0 \nonumber \\
\sum_k \sum_d 2(F^d_k)^T F^d_k v - 2(F^d_k)^T\textbf{a}^{m,d}_k  + (2v-2y) = 0 \nonumber \\
v = \bigg(\sum_k \sum_d (F^d_k)^T F^d_k + I\bigg)^{-1}  \bigg(\sum_k \sum_d(F^d_k)^T\textbf{a}^{m,d}_k + y\bigg) \nonumber 
\end{gather}

We can extract $\textbf{a}^m_*$ by reshaping v back into a matrix. 

\subsubsection{M-step}
\label{subsec:cont_mstep}

Given the point estimate for $\textbf{a}^m_*$, the log-likelihood Equation (\ref{continuousll}) can now be written as a function of the model parameters.
\begin{align}
\log \mathcal{L} = \sum_{m=1}^M \sum_{k=1}^K\log p(\textbf{a}^m_k | \textbf{a}^m_*; F^d_k, \tau_k) + \log p(\textbf{a}^m_*; \Theta, \sigma)  \nonumber 
\end{align}

In the M-step, we optimize the above equation with respect to the parameters $\Phi = \{F_k, \tau_k, \Theta,\sigma\}$. 
\begin{gather}
Q(F_k, \tau_k, \Theta, \sigma) = \sum_{m=1}^M \sum_{k=1}^{K}\log p(\textbf{a}^m_k | \textbf{a}^m_*;F^d_k, \tau_k) + \nonumber \\ \log p(\textbf{a}^m_*; \Theta, \sigma)
\label{contobj} 
\end{gather}

\textbf{Estimating $\bm{F^d_k}$:} Since each $F^d_k$ is a filter matrix constructed from a vector $f^d_k \in \rm I\!R^{WD}$, we differentiate \ref{contobj} with respect to $f^d_k$.
\begin{gather}
\Delta_{f^d_k} Q = 0 \nonumber \\
\Delta_{f^d_k} \sum_{m=1}^{M_k}\log p(\textbf{a}^m_k | \textbf{a}^m_*; F^d_k, \tau_k) = 0 \nonumber \\ 
\Delta_{f^d_k} \sum_{m=1}^{M_k}||\textbf{a}^{m,d}_k - F^d_k \text{vec}(\textbf{a}^m_*)||^2_2 = 0 \nonumber
\end{gather}

In the last step we make use of the fact that $\textbf{a}^m_k$ depends on $\textbf{a}^m_*$ through Gaussian noise. We also discard all other dimensions $d'\neq d $ since these do not depend on $f^d_k$. To estimate $f^d_k$, we can rearrange $F^d_k\text{vec}(\textbf{a}^m_*)$ such that $f^d_k$ is now the parameter vector of a linear regression problem with the independent variables represented by matrix $A$ which is obtained by creating a filtering matrix out of $\text{vec}(\textbf{a}^m_*)$. Hence, the optimization problem becomes
\begin{align}
\Delta_{f^d_k} \sum_{m=1}^{M_k} ||\textbf{a}^{m,d}_k - A f^d_k||^2_2 = 0 \nonumber \\
\therefore f^d_k = \bigg(\sum_{m=1}^{M_k} A^TA\bigg)^{-1}\bigg(\sum_{m=1}^{M_k} A^T\textbf{a}^{m,d}_k\bigg) \nonumber
\end{align}

\textbf{Estimating $\bm{\tau_k}$} Differentiating Equation (\ref{contobj}) with respect to $\tau_k$ and equating the gradient to 0, we have.

\begin{gather}
\Delta_{\tau_k} Q = 0 \nonumber \\
\Delta_{\tau_k} \sum_{m=1}^{M_k}\log p(\textbf{a}^m_k | \textbf{a}^m_*; F^d_k, \tau_k) = 0 \nonumber \\
\Delta_{\tau_k} \sum_{m=1}^{M_k} \sum_{d} \log \frac{1}{|2\pi\tau^2_kI|^\frac{1}{2}} e^{-\frac{1}{2\tau^2_k}||\textbf{a}^{m,d}_k - F^d_k\text{vec}(\textbf{a}^m_*)||^2_2} = 0 \nonumber \\
\Delta_{\tau_k} \sum_{m=1}^{M_k} \sum_{d} -T \log \tau_k - \frac{1}{2\tau_k^2} ||\textbf{a}^{m,d}_k - F^d_k\text{vec}(\textbf{a}^m_*)||^2_2 = 0 \nonumber \\
\frac{-M_kDT}{\tau_k} + \frac{1}{\tau_k^3} \sum_{m=1}^{M_k} \sum_{d} ||\textbf{a}^{m,d}_k - F^d_k\text{vec}(\textbf{a}^m_*)||^2_2 = 0 \nonumber \\
\tau^2_k = \frac{1}{M_kDT}\sum_{m=1}^{M_k} \sum_{d} ||\textbf{a}^{m,d}_k - F^d_k\text{vec}(\textbf{a}^m_*)||^2_2 \nonumber
\end{gather}

\textbf{Estimating $\bm{\Theta}$} Differentiating Equation (\ref{contobj}) with respect to $\Theta$ and equating the gradient to 0, we have.

\begin{gather}
\Delta_\Theta Q = 0 \nonumber \\
\Delta_\Theta \sum_{m=1}^M||\text{vec}(\textbf{a}^m_*) - \text{vec}(\textrm{X}_m\Theta)||^2_2 = 0 \nonumber
\end{gather}

By definition, each column of $\Theta$ is independent of each other. Hence we can estimate each $\theta^d$ separately (taking derivatives with respect to above equation would cancel all terms except those in $\theta^d$).

\begin{gather}
\Delta_{\theta^d} \sum_{m=1}^M (\textbf{a}^{m,d}_* - \textrm{X}_m\theta^d)^T(\textbf{a}^{m,d}_* - \textrm{X}_m\theta^d)= 0 \nonumber \\
\Delta_\Theta \sum_{m=1}^M (\textbf{a}^{m,d}_*)^T(\textbf{a}^{m,d}_*) - 2(\textbf{a}^{m,d}_*)^T\textrm{X}_m\theta^d+ (\theta^d)^T\textrm{X}_m^T\textrm{X}_m\theta^d = 0 \nonumber \\
\theta^d = \bigg(\sum_{m=1}^M \textrm{X}_m^T\textrm{X}_m\bigg)^{-1}\bigg(\sum_{m=1}^M\textrm{X}_m^T\textbf{a}^{m,d}_*\bigg) \nonumber 
\end{gather}

We can combine the estimation of all the columns of $\Theta$ as follows. 

\begin{gather}
\Theta = \bigg(\sum_{m=1}^M \textrm{X}_m^T\textrm{X}_m\bigg)^{-1}\bigg(\sum_{m=1}^M\textrm{X}_m^T\textbf{a}^m_*\bigg) \nonumber 
\end{gather}

\textbf{Estimating $\bm{\sigma}$} Differentiating Equation (\ref{contobj}) with respect to $\sigma$ and equating the gradient to 0, we have.

\begin{gather}
\Delta_{\sigma} Q = 0 \nonumber \\
\Delta_{\sigma} \sum_{m=1}^{M}\log p(\textbf{a}^m_*; \Theta, \sigma) = 0 \nonumber
\end{gather}

From Equation \ref{eqn:contastar}, $\textbf{a}^m_*$ was defined by adding zero mean Gaussian noise to $\text{vec}(\textbf{a}^m_*)$. Assuming $v=\text{vec}(\textbf{a}^m_k)$ and $y=\text{vec}(\textrm{X}_m\Theta)$, we have

\begin{gather}
\Delta_{\sigma} \sum_{m=1}^{M}\log \frac{1}{|2\pi\sigma^2I|^\frac{1}{2}} e^{-\frac{1}{2}(v - y)^T(\sigma^2I)^{-1}(v - y)} = 0 \nonumber \\
\Delta_{\sigma} \sum_{m=1}^{M} -TD \log \sigma - \frac{1}{2\sigma^2} ||v - y||^2_2 = 0 \nonumber \\
\sum_{m=1}^{M} \frac{-TD}{\sigma} + \frac{1}{\sigma^3} ||v - y||^2_2 = 0 \nonumber \\
\therefore \sigma^2 = 
\frac{1}{MTD}\sum_{m=1}^{M} ||\text{vec}(\textbf{a}^m_k) - \text{vec}(\textrm{X}_m\Theta)||^2_2 \nonumber
\end{gather}

\end{appendices}

\end{document}